\documentclass[acmsmall,pdftex]{acmart}

\usepackage{xcolor}
\usepackage{framed}
\usepackage{amsmath,amsfonts}
\usepackage{algorithmic}
\usepackage{epstopdf}
\usepackage{graphicx}
\usepackage{textcomp}
\usepackage{pifont}
\usepackage{makecell,multirow}
\usepackage{subfigure}
\usepackage{adjustbox}
\usepackage{caption}
\usepackage{balance}
\usepackage{url}
\usepackage{threeparttable}
\usepackage{tikz}
\usepackage{textcomp,booktabs}
\usepackage{diagbox}
\usepackage{colortbl}
\usepackage{tcolorbox}
\usepackage{enumitem}
\usepackage{bm}
\usepackage{fancyhdr}
\usepackage{soul}
\usepackage{listings}
\usepackage[normalem]{ulem}

\AtBeginDocument{%
  \providecommand\BibTeX{{%
    \normalfont B\kern-0.5em{\scshape i\kern-0.25em b}\kern-0.8em\TeX}}}

\setcopyright{acmcopyright}
\copyrightyear{2022}
\acmYear{2022}
\acmDOI{10.1145/1122445.1122456}

\acmJournal{TOSEM}
\acmVolume{37}
\acmNumber{4}
\acmArticle{111}
\acmMonth{1}

\begin{document}

\newcommand{\SpringGreen}[1]{\textcolor[RGB]{60,179,113}{#1}}
\newcommand{\RoyalBlue}[1]{\textcolor[RGB]{65,105,225}{#1}}
\newcommand{\DarkGreen}[1]{\textcolor[RGB]{0,100,0}{#1}}
\newcommand{\SlateBlue}[1]{\textcolor[RGB]{199,21,133}{#1}}
\newcommand{\yl}[1]{\SlateBlue{[Lin:#1]}}
\newcommand{\jj}[1]{\textcolor{orange}{[Junjie:#1]}}
\newcommand{\ky}[1]{ \SpringGreen{[KY:#1]}}
\newcommand{\gst}[1]{{\color{green}[GST:#1]}}
\newcommand{\gzh}[1]{\DarkGreen{[GZH:#1]}}
\newcommand{\lha}[1]{\RoyalBlue{[LHA:#1]}}

\newcommand{\imppca}{Improved PCA}
\newcommand{\tech}{\textit{SemPCA}}
\newcommand{\RQOne}{Can \tech{} achieve comparable effectiveness with the existing log-based anomaly detection techniques (especially the advanced DL-based techniques)?}
\newcommand{\RQTwo}{Can \tech{} perform more stably under insufficient training data than the existing techniques?}
\newcommand{\RQThree}{How does \tech{} perform in terms of efficiency compared with the existing techniques?}
\newcommand{\RQFour}{Is \tech{} interpretable for the detected anomalies?}
\newcommand{\RQFive}{What happens when paring is incomplete?}

\newcommand{\HDFS}{\textit{HDFS}}
\newcommand{\BGL}{\textit{BGL}}
\newcommand{\Spirit}{\textit{Spirit}}
\newcommand{\NC}{\textit{NC}}
\newcommand{\ZQY}{\textit{MC}}

\newcommand{\LAD}{log-based anomaly detection}

\newcommand{\finding}[1]{ \begin{tcolorbox}
\textbf{Finding \refstepcounter{num}\thenum}: #1
\end{tcolorbox}}

\newcommand{\del}[1]{}
\newcommand{\ins}[1]{{\color{black}{#1}}}

\lstdefinestyle{myCustomStyle}{
    backgroundcolor=\color{white},   
    basicstyle=\ttfamily\tiny\color{black},
    keywordstyle=\color{black},      
    commentstyle=\color{black},      
    stringstyle=\color{black},       
    breakatwhitespace=false,         
    breaklines=true,                 
    captionpos=b,                    
    keepspaces=true,                 
    numbers=left,                    
    showspaces=false,                
    showstringspaces=false,
    showtabs=false,                  
    tabsize=2,
    frame=single,                    
    framesep=2pt,         
}
\renewcommand{\lstlistingname}{Log-Seq}
\title{Try with Simpler -- An Evaluation of Improved Principal Component Analysis in Log-based Anomaly Detection}

\author{Lin Yang}
\email{linyang@tju.edu.cn}
\affiliation{%
  \institution{College of Intelligence and Computing, Tianjin University}
  \streetaddress{135th, Yaguan Road}
  \city{Tianjin}
  \state{Tianjin}
  \country{China}
  \postcode{30072}
}

\author{Junjie Chen}
\authornote{Corresponding author.}
\affiliation{%
  \institution{College of Intelligence and Computing, Tianjin University}
  \country{China}
}
\email{junjiechen@tju.edu.cn}

\author{Shutao Gao}
\affiliation{%
  \institution{Tianjin International Engineering Institute, the School of Future Technology, Tianjin University}
  \country{China}
}
\email{gaoshutao@tju.edu.cn}

\author{Zhihao Gong}
\affiliation{%
  \institution{College of Intelligence and Computing, Tianjin University}
  \country{China}
}
\email{gongzhihao@tju.edu.cn}

\author{Hongyu Zhang}
\affiliation{%
  \institution{Chongqing University}
  \country{China}
}
\email{hyzhang@cqu.edu.cn}

\author{Yue Kang}
\affiliation{%
  \institution{College of Intelligence and Computing, Tianjin University}
  \country{China}
}
\email{kangyue0327@tju.edu.cn}

\author{Huaan Li}
\affiliation{%
  \institution{College of Intelligence and Computing, Tianjin University}
  \country{China}
}
\email{devon_zw@tju.edu.cn}

\renewcommand{\shortauthors}{Lin Yang and Junjie Chen, et al.}

\begin{abstract}
With the rapid development of deep learning (DL), the recent trend of log-based anomaly detection focuses on extracting semantic information from log events (i.e., templates of log messages) and designing more advanced DL models for anomaly detection.
Indeed, the effectiveness of log-based anomaly detection can be improved, but these DL-based techniques further suffer from the limitations of  heavier dependency on training data (such as data quality or data labels) and higher costs in time and resources due to the complexity and scale of DL models, which hinder their practical use.
On the contrary, the techniques based on traditional machine learning or data mining algorithms are less dependent on training data and more efficient, but produce worse effectiveness than DL-based techniques which is mainly caused by the problem of unseen log events (some log events in incoming log messages are unseen in training data) confirmed by our motivating study.
Intuitively, if we can improve the effectiveness of traditional techniques to be comparable with advanced DL-based techniques, log-based anomaly detection can be more practical.
Indeed, an existing study in the other area (i.e., linking questions posted on Stack Overflow) has pointed out that traditional techniques with some optimizations can indeed achieve comparable effectiveness with the state-of-the-art DL-based technique, indicating the feasibility of enhancing traditional log-based anomaly detection techniques to some degree.

Inspired by the idea of ``try-with-simpler'', we conducted the first empirical study to explore the potential of improving traditional techniques for more practical log-based anomaly detection.
In this work, we optimized the traditional unsupervised PCA (Principal Component Analysis) technique by incorporating a lightweight semantic-based log representation in it, called \tech{}, and conducted an extensive study to investigate the potential of \tech{} for more practical log-based anomaly detection.
By comparing seven log-based anomaly detection techniques (including four DL-based techniques, two traditional techniques, and \tech{}) on both public and industrial datasets, our results show that \tech{} achieves comparable effectiveness as advanced supervised/semi-supervised DL-based techniques while being much more stable under insufficient training data and more efficient,
demonstrating that the traditional technique can still excel after small but useful adaptation. \end{abstract}



\maketitle
\section{Introduction}
\label{sec:introduction}
With the scale and complexity of software increasing, logs have become more important for software maintenance~\cite{DBLP:series/synthesis/2017Yao,DBLP:conf/icse/0003HLXZHGXDZ19,breier2015anomaly,DBLP:conf/srds/Zhang0MZ20,icse16_logging_practice,DBLP:journals/infsof/WangCYZW23,DBLP:conf/kbse/0003HL0HGXDZ19}.
Logs record system events and states of interest and are produced during system runtime.
Developers can check software status, detect anomalies, and diagnose root causes by carefully inspecting the recorded logs.
However, log data tend to be massive due to the large software scale~\cite{logzip,BGL,Spirit}, and thus manual inspection for logs is very difficult or even infeasible.
Therefore, over the years, a large amount of work on automated log analysis has been conducted~\cite{LogAnalysisServeyCUHK,DBLP:conf/issre/HeZHL16,DBLP:journals/csur/HeHCYSL21,fse21nengwen,DBLP:conf/icse/ZhuHLHXZL19,DBLP:conf/icse/ZhuHFZLZ15,LogLens,microserviceanomaly_pengxin,DBLP:conf/sigsoft/Zhao0YWLQXZSP21,DBLP:conf/iwqos/Li0JZWZWJYWCZNS21,DBLP:conf/icws/WangZCLZS20,DBLP:journals/tosem/ChenCWZWCZW23}.

Log-based anomaly detection is one of the most important tasks in automated log analysis, which aims to automatically detect software anomalies in time based on log data so as to reduce the loss caused by these anomalies~\cite{PCA,LogCluster,SVM,DecisionTree,InvariantMining,DeepLog,LogAnomaly,PLELog,NeuralLog,LADCNN,lstm_based_anomaly_detection}.
The general process of log-based anomaly detection consists of two main steps: representing log events (i.e., templates of log messages) and log sequences (i.e., series of log events that record specific execution flows) as vectors, and then building an anomaly detection model via machine learning (ML) or data mining (DM) algorithms based on those log vectors.
With the rapid development of deep learning (DL), the trend of log-based anomaly detection focuses on two aspects in recent years~\cite{LADCNN,DeepLog,LogAnomaly,PLELog,NeuralLog,LogGAN,Trine}:
1) more effectively extracting semantic information from log events for log representation, and 2) designing more advanced DL models for anomaly detection.
The first aspect can help understand the semantics of log events, which facilitates to relieve the problem of unseen log events.
This problem is mainly caused by software evolution (e.g., frequent modification of log statements in source code), resulting in that some incoming log events may not appear in training data~\cite{LogRobust,PLELog}.
It is harmful to the effectiveness of log-based anomaly detection as demonstrated in the existing work~\cite{LogRobust,LogAnomaly,PLELog}.
The second aspect aims to build a more accurate model to learn the distinction between anomalies and normal cases. 

Although the effectiveness of log-based anomaly detection is indeed improved, these advanced DL-based techniques further suffer from the limitations shared in the area of DL, which can largely hinder their practical use.
First, they heavily rely on training data (such as data quality or data labels).
This is because building more complex DL models has a higher requirement for the volume and distribution of training data (which has been confirmed by our study in Section~\ref{sec:motivating_study}.)
Besides, supervised DL algorithms also rely on data labels.
In practice, it is very hard to collect sufficient high-quality data with labels since manual labeling is costly (especially for tasks that require domain knowledge). Furthermore, new data has to be frequently collected due to frequent software evolution, which further aggravates the difficulty of training data collection.
Second, these DL-based techniques tend to spend much time and computing resources on hyper-parameter tuning, model building, and prediction due to the complexity of DL models and the scale of training data.
However, in practice, both efficient deployment of accurate anomaly detection models and real-time anomaly detection are desired~\cite{empirical_efficiency}.

On the contrary, traditional techniques adopt lightweight ML or DM algorithms (e.g., clustering algorithms~\cite{LogCluster} and PCA -- Principal Component Analysis~\cite{PCA}) to build anomaly detection models.
Hence, they are more efficient and less dependent on training data, especially for unsupervised algorithms (e.g., PCA~\cite{PCA}).
However, as demonstrated in an existing study~\cite{DBLP:conf/issre/HeZHL16}, these traditional log-based anomaly detection techniques have lower accuracy than the advanced DL-based techniques, which limits their practical use. 
Moreover, one major reason for the poor effectiveness of traditional techniques lies in that they do not consider the influence of unseen log events~\cite{PLELog}, which is also confirmed by our motivating study (to be presented in Section~\ref{sec:motivating_study}).
Intuitively, if we can improve the effectiveness of traditional techniques and make them comparable with advanced DL-based ones in terms of effectiveness, log-based anomaly detection can be more practical.
Previously, Fu et al.~\cite{menzies_easy_over_hard} proposed the idea of ``try-with-simpler'', which pointed out that for the task of linking questions posted on Stack Overflow, traditional SVM (Support Vector Machine) with some optimizations can achieve comparable effectiveness with the state-of-the-art DL-based technique.
This also indicates the feasibility of enhancing traditional log-based anomaly detection techniques to some degree.
However, this direction remains unexplored in the area of log-based anomaly detection.

Inspired by the idea of ``try-with-simpler''~\cite{menzies_easy_over_hard}, we conducted the first empirical study to explore the potential of improving traditional techniques for more practical log-based anomaly detection.
Here, we studied the typical traditional technique (i.e., PCA~\cite{PCA}) and optimized it by incorporating semantic-based log representation.
For ease of presentation, we call the optimized PCA technique \tech{}.
This is because (1) PCA is the most widely-studied \textit{unsupervised} traditional technique in the literature~\cite{DBLP:journals/corr/abs-1909-03495,DBLP:journals/ijcse/CallegariDGP18,DBLP:journals/corr/abs-1801-01571,rao1964use,abdi2010principal,shyu2003novel} (unsupervised algorithms tend to be more practical in terms of efficiency and data dependency~\cite{advantage_of_unsupervised});
(2) Ignoring the influence of unseen log events is a major cause to the poor effectiveness of traditional techniques, which can be relieved by extracting semantic information from log events for log representation~\cite{LogRobust,PLELog}.
According to the existing studies~\cite{LogAnomaly,LogRobust,PLELog,NeuralLog}, 
the above-mentioned two limitations (i.e., depending on training data heavier and spending more cost on model construction) are mainly caused by the advanced DL models rather than the extraction of semantic information from log events.
Therefore, incorporating semantic-based log representation into the traditional PCA technique does not make \tech{} further suffer from the two limitations.
Specifically, we incorporated a simple and lightweight semantic-based log representation method into the traditional PCA technique in order to avoid incurring too much extra cost, which aggregates the word vectors in each log event by measuring the importance of each word through TF-IDF (Term Frequency - Inverse Document Frequency)~\cite{tfidf98,tfidf00}.

Based on \tech{}, our empirical study addressed three research questions to investigate whether improving the traditional technique with simple adaptation can make log-based anomaly detection more practical.
\begin{itemize}
    \item \textbf{RQ1}: \RQOne{}
    \item \textbf{RQ2}: \RQTwo{}
    \item \textbf{RQ3}: \RQThree{}
\end{itemize}
\noindent In the study, we compared \tech{} with a set of widely-studied log-based anomaly detection techniques (including four DL-based ones and two traditional ones) on five datasets (including three widely-used public datasets -- \HDFS{}~\cite{HDFS}, \BGL{}~\cite{BGL}, and \Spirit{}~\cite{Spirit}, and two industrial datasets).
The two industrial datasets are collected from two real-world systems in two organizations (i.e., the network center of our university and one influential motor corporation throughout the world\footnote{We hide the company name due to the company policy.}), since \tech{} has been successfully applied to them due to its practical value.

Our results show that the optimized version of the unsupervised PCA technique (i.e., \tech{}) achieves comparable effectiveness with supervised/semi-supervised DL-based techniques, e.g., the average F1-score of \tech{} across all datasets is 0.959 while that of the most effective DL-based technique (i.e., the supervised LogRobust~\cite{LogRobust}) is 0.983.
Moreover, \tech{} has been demonstrated to be much more efficient for model building and prediction, and more stable when facing insufficient training data.
For example, for model building, \tech{} is at least 5,808X faster than the advanced DL-based techniques (e.g., LogRobust~\cite{LogRobust}).
The results demonstrate the significant practicability of \tech{}, indicating that the typical traditional PCA technique can still excel in log-based anomaly detection after simple adaptation, also confirming the idea of ``try-with-simpler'' in our task.

To sum up, our work makes the following major contributions:

\begin{itemize}
    \item We conduct a dedicated study, as the first attempt, to investigate whether traditional techniques through some optimizations can achieve comparable effectiveness with state-of-the-art DL-based techniques in log-based anomaly detection. 

    \item We improve the traditional PCA technique by incorporating a lightweight semantic-based log representation method.
    By comparing with a set of advanced techniques on five datasets, it achieves comparable effectiveness while is more efficient and stable.
    
    
    \item We develop and release a toolbox that integrates the implementations of all the studied log-based anomaly detection techniques, in order to promote future research, practical use, and evaluation replication.

\end{itemize}

\section{Background}
\label{sec:background}

\subsection{Log Terminology}
\label{sec:related_terminologies}
\begin{figure}
    \centering
    \includegraphics[scale=0.45, bb=0 0 780 120]{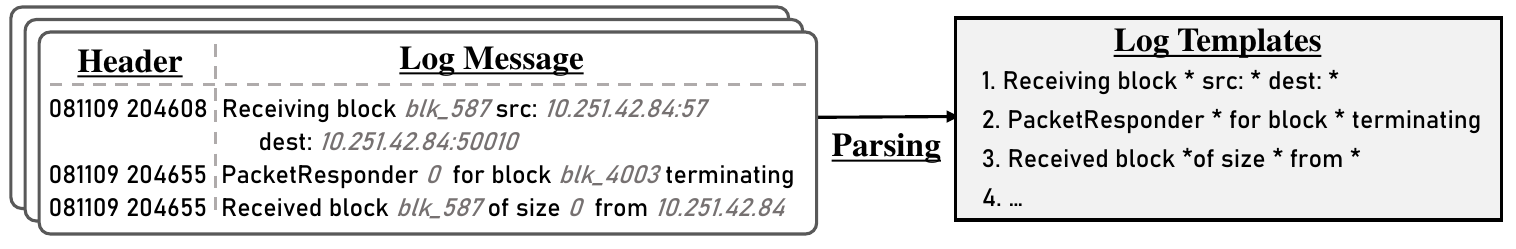}
    \caption{An illustrating example for log terminology}
    \label{fig:log_term}
\end{figure}

Figure~\ref{fig:log_term} shows an example of log data sampled from the public HDFS dataset~\cite{HDFS}, facilitating the illustration of log terminology.
From this figure, a \textbf{log message} is a raw unstructured sentence generated during system runtime, which records software status of the time.
A log message (e.g., \textit{PacketResponder 0 for block blk\_4003 terminating}) consists of a \textbf{log event} (e.g., \textit{PacketResponder * for block * terminating}) and \textbf{log parameters} (e.g., \textit{0, blk\_4003}).
The former is the template of a log message, which is programmed by developers as a log statement in source code.
This is a constant part in a log message.
The latter is a variable part that records some software attributes, such as IP address, file path, and block ID.
In the area of log-based anomaly detection, log events can be automatically extracted from log messages via \textbf{log parsing}, which is the first step in log-based anomaly detection and has been widely studied over the years~\cite{DBLP:conf/icse/ZhuHLHXZL19,Drain,Logram,SLCT,LogCluster_Parser,DBLP:conf/msr/NagappanV10,Lognroll}.

In practice, developers as well as log-based anomaly detection techniques tend to understand a sequence of log messages for checking software status.
Here, we call a sequence of log events \textbf{log sequence}, which usually reflects a specific execution flow or the system status within a certain period.
By checking the task ID of each log message or adopting some strategies (e.g., splitting windows), we can know which log events belong to the same log sequence. 
Based on whether or not the log sequences indicate software anomalies, they can be classified as \textbf{anomalous log sequences} and \textbf{normal log sequences}.

\subsection{Log-based Anomaly Detection Techniques}
\label{sec:studied_techniques}


Due to the importance of log-based anomaly detection, a lot of techniques have been proposed in the literature~\cite{LogCluster,PCA,DeepLog,LogAnomaly,LogRobust,PLELog}. 
As presented in Section~\ref{sec:introduction}, these techniques share the same high-level process: 1) representing log events and log sequences as vectors, and 2) building an anomaly detection model through traditional or deep learning algorithms based on log vectors.
With the rapid development of DL, the recent trend {of}\del{on} log-based anomaly detection is to incorporate more advanced DL algorithms in order to improve the anomaly detection effectiveness.
Here, according to whether they are based on traditional ML/DM algorithms or advanced DL algorithms, we classify them into two categories: \textit{traditional techniques} and \textit{DL-based techniques}.
In the following, we introduce two traditional techniques and four DL-based techniques in detail because they are the most widely-studied in the existing studies~\cite{DBLP:conf/issre/HeZHL16,LogAnomaly,PLELog}.
Following these existing work, we also used them as the studied techniques in our study. 

\subsubsection{Traditional Log-based Anomaly Detection Techniques}
\label{sec:traditional}
The two most widely-studied traditional log-based anomaly detection techniques are PCA~\cite{PCA} and LogCluster~\cite{LogCluster}. 

\textbf{PCA} is one of the unsupervised techniques among these widely-studied techniques. 
It first represents each log sequence as an \textit{Event Count Vector} by counting the occurrence of each log event in the log sequence.
Then, it applies the PCA algorithm~\cite{PrincipleComponentAnalysis,abdi2010principal} 
to identify principal components (which preserve data variation as much as possible) from event count vectors and projects those vectors to the space of principal components (called normal space). 
Based on the projections to the normal space, the projection of each vector to the abnormal space can be obtained accordingly.
Since the projections of normal log sequences to the abnormal space should be less discrete, the anomalies can be detected through outlier detection in the abnormal space. 
Specifically, if the Square Prediction Error (SPE)~\cite{q_static} of a projection to the abnormal space is larger than a given threshold, the log sequence is regarded as an anomalous one.

\textbf{LogCluster} is a semi-automatic and supervised technique initially.
It clusters training data into several groups, each of which is manually labeled as a normal or anomalous group.
By determining which kind of groups an incoming log sequence belongs to, the anomaly detection result can be obtained.
If an incoming log sequence does not belong to any of these groups, developers have to manually check it.
To make LogCluster easy-to-study in practice, an adapted version of LogCluster is used in many existing studies~\cite{DBLP:conf/issre/HeZHL16,LogAnomaly,NeuralLog,PLELog} (as well as our study), which is an automatic and semi-supervised technique.
Specifically, it labels and clusters only normal training data to obtain normal groups.
If an incoming log sequence does not belong to any of the groups, it is regarded as an anomaly.
Before clustering through the Agglomerative Hierarchical Clustering algorithm~\cite{hierarchical_clustering}, it first calculates the weight of each log event by an IDF-based weighting~\cite{DBLP:books/daglib/0021593} and a contrast-based weighting~\cite{LogCluster}, and then a log sequence is represented as a \textit{Weighted Event Count Vector} based on the weights of the log events in the log sequence.

\subsubsection{DL-based Log-based Anomaly Detection Techniques}
\label{sec:DL}
The four most widely-studied DL-based techniques are DeepLog~\cite{DeepLog}, LogRobust~\cite{LogRobust}, LogAnomaly~\cite{LogAnomaly}, and PLELog~\cite{PLELog}.

\textbf{DeepLog} is a widely-studied DL-based technique, which learn log patterns from the system's normal executions to build the anomaly detection model~\cite{DeepLog}.
It first represents each log event as a unique ID and thus each log sequence is represented as a sequence of IDs (also called \textit{Log Event ID Sequence}). 
Then it builds an LSTM (Long Short Term Memory~\cite{LSTM_in_NLP}) model that can predict the next ID according to the sequence of IDs occurring before the one to be predicted in a log sequence.
If an ID is mispredicted in an incoming log sequence, this log sequence is regarded as an anomaly.

\textbf{LogRobust}~\cite{LogRobust} is a state-of-the-art supervised log-based anomaly detection technique. 
It is also the first to relieve the problem of unseen log events in log-based anomaly detection.
It proposes the \textit{TF-IDF-based Semantic Representation} method to represent each log event, which first adopts the pre-trained word vectors on the Common Crawl Corpus dataset using the Fast-Text toolkit~\cite{bojanowski2017enriching,joulin2017bag,joulin2016fasttext} to represent each word in a log event as a vector, and then aggregates these word vectors as a log-event vector by measuring the importance of each word through TF-IDF~\cite{tfidf98,tfidf00}.
Based on these log-event vectors from training data, LogRobust builds a supervised model through attention-based Bi-directional LSTM~\cite{LSTM_in_NLP}, which can predict whether an incoming log sequence is normal or anomalous.

\textbf{LogAnomaly} extends DeepLog by improving log representation, which combines both \textit{Event Count Vector} and \textit{Synonyms-Antonyms-based Semantic Representation}~\cite{Log2Vec}.
The insight of synonyms-antonyms-based representation is to capture the semantic information of the involved synonyms and antonyms when representing a log event.
This is because similar log events may also have very different semantics, e.g., ``Interface * changed state to \textit{\textbf{down}}'' and ``Interface * changed state to \textit{\textbf{up}}'', and capturing the semantic information of synonyms and antonyms can help relieve this issue to obtain better log representation.
This method requires to construct the set of synonyms and antonyms with the aid of operators based on domain knowledge.
Then, it builds an embedding model based on dLCE~\cite{dLCE} to represent a log event as a semantic vector.
Based on the log representation, LogAnomaly learns both semantic patterns and quantity patterns from the system's normal execution by two separate LSTM models.
Similar to DeepLog, if a log event is mispredicted based on the log events occurring before this event in an incoming log sequence, this log sequence is regarded as an anomaly.


\textbf{PLELog} is the state-of-the-art semi-supervised log-based anomaly detection technique, which requires a set of labeled normal training data among the whole training set~\cite{PLELog}.
Its main contribution is to estimate the labels of the remaining unlabeled training data by clustering all the training data through the \textit{HDBSCAN} algorithm~\cite{HDBSCAN}.
According to the clustering results and the known normal log sequences, PLELog assigns a probabilistic label to each unlabeled log sequence in training data. 
PLELog also uses the \textit{TF-IDF-based Semantic Representation} method to represent each log event for relieving the problem of unseen log events.
In particular, it adopts GloVe, a pre-trained language model based on global vectors~\cite{pennington2014glove}, to represent each word in a log event.
Based on these vectors and estimated labels, it builds an model through attention-based Bi-directional GRU~\cite{compare_gru_lstm}, which can predict whether an incoming log sequence is normal or anomalous.

\subsubsection{Summary}
\label{sec:techsummary}
For ease of understanding, Table~\ref{tab:studied_approach_base} briefly summarizes the log-representation method and the algorithm for building the anomaly detection model used by each technique.
In fact, both semantic-based log representation and advanced DL models contribute to the superiority of DL-based techniques over traditional techniques in terms of the anomaly detection effectiveness.
Specifically, the former improves the effectiveness by relieving the problem of unseen log events as presented in Section~\ref{sec:introduction}.
Moreover, it does not incur the above-mentioned limitations (i.e., depending on training data more heavily and spending more time on model construction) for DL-based techniques, since these semantic-based log representation methods directly use pre-trained language models to extract semantic information from log events~\cite{NeuralLog,DBLP:journals/corr/abs-2102-11570}.
In other words, the above-mentioned limitations are mainly caused by the latter (i.e., advanced DL models), but the effectiveness improvement is brought by the two aspects.
Therefore, this motivates a potential direction to optimize traditional techniques by incorporating semantic-based log representation, which is likely to improve the anomaly detection effectiveness but does not damage the advantages of high efficiency and weak data dependency.


\useunder{\uline}{\ul}{}
\begin{table}[t]
\caption{Summary of the Studied Log-based Anomaly Detection Techniques}
\label{tab:studied_approach_base}
\centering
\begin{tabular}{@{}c|l|l|l@{}}
\toprule
\multicolumn{1}{c|}{\textbf{Category}}     & \textbf{Technique}                   & \textbf{Log Representation}          & \textbf{Anomaly Detection}      \\ \midrule
\multirow{2}{*}{Traditional} & PCA                         & Event Count Vector          & PCA                    \\ \cmidrule{2-4} 
                             & LogCluster                  & Weighted Event Count Vector & Hierarchal Clustering  \\ \midrule
\multirow{6}{*}{DL-based}    & DeepLog                     & Log Event ID Sequence       & LSTM                   \\ \cmidrule{2-4}  
                             & LogAnomaly & \begin{tabular}[c]{@{}l@{}}Event Count Vector\& \\ Synonyms-Antonyms-based \\     Semantic Representation\end{tabular}      & Attention-based Bi-LSTM  \\
                             \cmidrule{2-4} 
 & LogRobust               & \begin{tabular}[c]{@{}l@{}}TF-IDF-based \\ Semantic Representation\end{tabular}                  & LSTM    \\ \cmidrule{2-4}  
 & \multirow{2}{*}{PLELog} & \multirow{2}{*}{\begin{tabular}[c]{@{}l@{}}TF-IDF-based \\ Semantic Representation\end{tabular}} & HDBSCAN\& \\
                             &                             &                             & Attention-based Bi-GRU \\ \bottomrule
\end{tabular}%
\end{table}

\section{Motivating Study}
\label{sec:motivating_study}
\newcounter{num}
In this section, we conducted a motivating study to quantitatively investigate the advantages and disadvantages of traditional techniques and DL-based techniques (mentioned in Section~\ref{sec:introduction}), which aims to motivate the potential of traditional techniques in log-based anomaly detection.
We first compared DL-based techniques and traditional techniques in terms of effectiveness, efficiency, and stability (Section~\ref{sec:rq1-1}).
As demonstrated in the existing study~\cite{le21howfar,menzies_easy_over_hard}, they are important factors affecting the practical use of log-based anomaly detection techniques.
Then, we investigated whether the problem of unseen log events is an important factor affecting the effectiveness of traditional techniques (Section~\ref{sec:rq1-2}).
Specifically, we used a state-of-the-art \textit{supervised} technique, i.e., LogRobust~\cite{LogRobust}, as the representative of DL-based techniques, and the \textit{unsupervised} technique, i.e., PCA~\cite{PCA}, as the representative of traditional techniques.
They are significantly different, and such a strong contrast is more helpful in motivating our work.
Here, we studied them on the public \Spirit{} dataset~\cite{Spirit}, which was produced by the Spirit supercomputer at Sandia National Labs with a time span of 2.5 years.
More details about this dataset can be found in Section~\ref{sec:dataset}).

\subsection{LogRobust v.s. PCA}
\label{sec:rq1-1}
We evaluated the \textit{effectiveness} of LogRobust and PCA on different sets of training data, which can help investigate their \textit{stability}.
Specifically, we split \Spirit{} into training data, validation data, and test data in chronological order of log sequences with the ratio of 6:1:3.
This can ensure that all log sequences in training data are produced before the log sequences in test data.
In this study, we kept the validation and test data unchanged and constructed several training sets by randomly sampling 20\% or 40\% log data from the whole training data.
We repeated the sampling process 10 times for each sampling ratio. 
In total, we obtained 20 training sets, i.e., 10 training sets with 20\% of training data and 10 training sets with 40\% of training data.
Then, we built an anomaly detection model based on each training set using LogRobust and PCA respectively, and measured their effectiveness on the test data in terms of F1-score~\cite{fscoreandroc}.
The hyper-parameter tuning process for each technique in each training set is conducted via grid search on the same validation data following the practice in DL~\cite{ripley2007pattern}.

\begin{figure}
    \centering
    \includegraphics[width=0.8\linewidth]{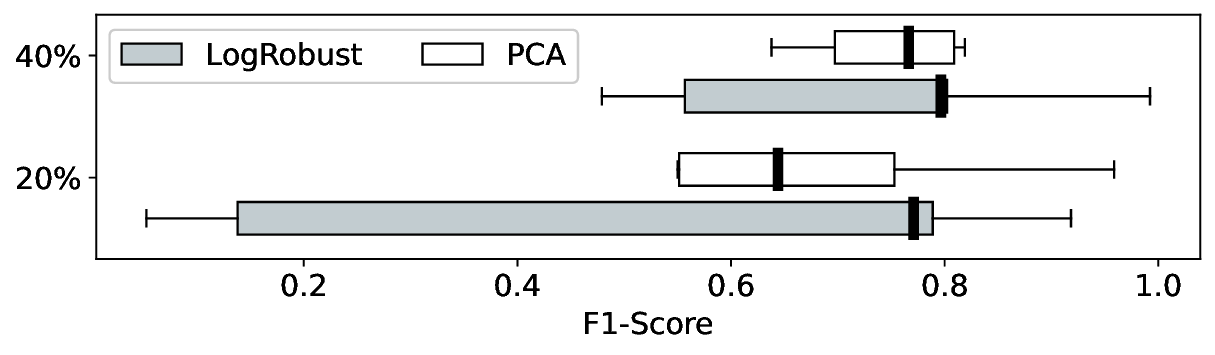}
    \vspace{-2mm}
    \caption{F1-score under different sets of training data}
    \vspace{-2mm}
    \label{fig:motivating_box}
\end{figure}

Figure~\ref{fig:motivating_box} shows F1-score of LogRobust and PCA, where each box shows the overall effectiveness of each technique across the 10 training sets with 20\% or 40\% of training data.
From this figure, the medium F1-score of LogRobust is larger than that of PCA regardless of 20\% or 40\% of training data, demonstrating the effectiveness of such an advanced supervised DL-based technique.
However, LogRobust performs more unstably than PCA under each studied amount of training data, indicating that the former relies more heavily on the training data.
Also, LogRobust becomes more stable with increasing amount of training data (from 20\% to 40\% in the study), indicating its heavy dependency on the training data volume.
However, in practice, collecting sufficient high-quality training data is very challenging and costly, which largely hinders the practicability of such a DL-based technique.

We further measured the \textit{time cost} spent on building an anomaly detection model based on the whole training data and predicting one incoming log sequence for LogRobust and PCA, respectively.
The time cost spent on training LogRobust is 2,361.429 seconds, while that of PCA is only 0.983 second.
Regarding the time cost spent on predicting incoming log sequences in the test data, LogRobust spends 1.704 seconds while PCA spends only 0.303 second.
The results demonstrate the significant superiority of PCA over LogRobust in terms of training and prediction efficiency.

\finding{The supervised DL-based technique (i.e., LogRobust) outperforms the unsupervised traditional technique (i.e., PCA) in terms of effectiveness (the medium F1-score). However, the latter is much more efficient and stable.}

\subsection{Influence of unseen log events on PCA}
\label{sec:rq1-2}
According to Finding 1, if we can improve the effectiveness of PCA without obviously damaging its efficiency and stability, more practical log-based anomaly detection can be achieved.
Hence, it is important to identify the important factor affecting the effectiveness of PCA.
As demonstrated in the existing studies~\cite{LogRobust,LogAnomaly,PLELog,DeepLog}, the problem of unseen log events is harmful to the effectiveness of log-based anomaly detection.
Here, we conducted an experiment to investigate whether this problem is also an important factor that negatively affects the effectiveness of PCA. 
Specifically, we constructed three training sets by randomly sampling 20\% training data from the whole training data.
In particular, we ensured that the three training sets have different numbers of unseen log events with regard to the test data.

Based on each training set, we built an anomaly detection model using PCA and measured its F1-score on the same test data. 
The results are shown in Table~\ref{tab:motivating_pca}. 
We found that with the number of unseen log events decreasing, the effectiveness of PCA becomes better in terms of F1-score, confirming the significant influence of the problem of unseen log events on PCA.
With this finding, the effectiveness of PCA could be effectively improved by relieving the problem of unseen log events. 
This further motivates the potential of such an unsupervised traditional technique for more practical log-based anomaly detection.

\begin{table}[t]
    \centering
    \caption{F1-score of PCA under different numbers of unseen log events}
    \vspace{-1mm}
    \begin{tabular}{@{}lrrr@{}}
    \toprule 
    \#Unseen Events            & 404  &  311   & 283  \\ \midrule
    \multicolumn{1}{r}{F1-score} & 0.738 & 0.872 & 0.935 \\ \bottomrule
    \end{tabular}
    \label{tab:motivating_pca}
\end{table}

\finding{Relieving the problem of unseen log events may be a promising direction to improve the effectiveness of PCA.}

\section{METHODOLOGY}
\label{sec:methodology}

Based on the findings of the motivating study, we conducted an extensive study, as the first attempt, to explore the potential of traditional techniques for more practical log-based anomaly detection. 
In this section, we present the methodology of our empirical study, including the optimized version of the traditional PCA technique and the study design.

\subsection{Studied techniques}
\label{sec:pca++}
As presented in Section~\ref{sec:studied_techniques}, we evaluated six widely-studied log-based anomaly detection techniques, including two traditional techniques (i.e., PCA and LogCluster) and four DL-based techniques (i.e., DeepLog, LogAnomaly, LogRobust, and PLELog).
Besides, the goal of our study is to explore whether traditional techniques through some adaptations can achieve comparable effectiveness with state-of-the-art DL-based techniques and thus make log-based anomaly detection more practical.
Therefore, motivated by the findings in our motivating study in Section~\ref{sec:motivating_study}, we optimized the traditional PCA technique by incorporating semantic-based log representation.
Please note that our work does not aim to propose a new technique, but explores the potential of existing traditional techniques through simple adaptations.
Hence, we directly adopted the existing semantic-based log representation method for optimizing the traditional PCA technique.
From Table~\ref{tab:studied_approach_base}, there are two semantic-based log representation methods (i.e., synonyms-antonyms-based~\cite{LogAnomaly} and TF-IDF-based semantic representation~\cite{LogRobust}) used in these widely-studied log-based anomaly detection techniques.
Following the idea of ``try-with-simpler'', we selected the lightweight TF-IDF-based semantic representation method (instead of Synonyms-Antonyms-based Semantic Representation) for optimizing the traditional PCA technique, since it is more light-weight without operator intervention and additional embedding model training.
For ease of presentation, we call the optimized version of PCA \tech{}.
In total, we studied seven log-based anomaly detection techniques in our study.

\begin{figure}[t]
    \centering
    \includegraphics[scale=0.71, bb=0 0 780 120]{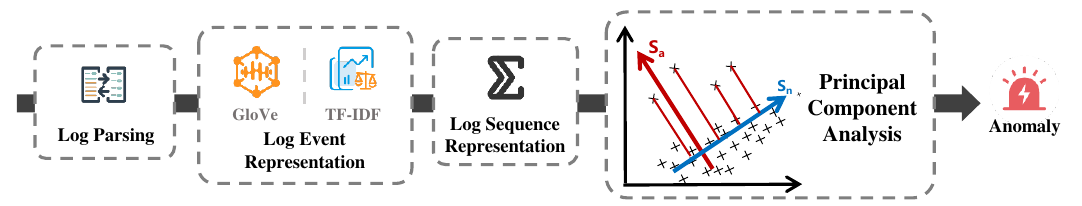}
    \caption{Workflow of \tech{}}
    \label{fig:overview}
\end{figure}

As the first attempt to explore the potential of traditional techniques for more practical log-based anomaly detection, we adopted the traditional PCA technique as the representative since it is a typical \textit{unsupervised} technique and unsupervised algorithms tend to be more practical in terms of efficiency and data dependency~\cite{advantage_of_unsupervised}.
Actually, other traditional techniques could be also combined with the semantic-based log representation method and thus achieve better effectiveness, which will be discussed in detail in Section~\ref{sec:other}.
Since the former six techniques have been presented before, we introduce \tech{} in detail here.
Figure~\ref{fig:overview} shows the workflow of \tech{}, also following the two main steps (i.e., log representation and anomaly detection model building).

\subsubsection{Semantic-based Log Representation}
\label{sec:pcarep}
\tech{} first extracts log events from log messages through log parsing and then represents log events as vectors.
Specifically, \tech{} adopts the widely-used log parser (i.e., Drain~\cite{Drain}) for log event extraction.
For fair comparison, all the studied log-based anomaly detection techniques use the same log parser in our study, to eliminate the influence of different log parsers on anomaly detection.

Afterwards, \tech{} adopts the lightweight TF-IDF-based semantic representation method to represent each log event.
In particular, \tech{} directly adopts the pre-trained word vectors on the Common Crawl Corpus dataset using the Fast-Text~\cite{joulin2016fasttext} toolkit for TF-IDF-based representation following the existing work~\cite{LogRobust,PLELog}.
Specifically, the representation of a log event (denoted as $e$) is calculated by $V(e) = \sum_{i=1}^{L_e} (\omega_i \times \alpha_i)$, where $\omega_i$ is the word vector of the $i^{\textit{th}}$ word (denoted as $w_i$) in $e$, $L_e$ is the number of words in $e$, and $\alpha_i =\textit{TF}(w_i) \times \textit{IDF}(w_i)$.
Here, following the existing work~\cite{PLELog}, we adopts GloVe, a pre-trained language model based on global vectors~\cite{pennington2014glove}, to represent each word in a log event as a word vector.
\textit{TF} (Term Frequency) measures how frequently a word $w$ occurs in $e$, i.e., $\textit{TF}(w) = \frac{\#w}{L_e}$, where $\#w$ is the number of the occurrence of $w$ in $e$.
\textit{IDF} (Inverse Document Frequency) measures how common or rare a word $w$ is in all the log events, i.e., $IDF(w) = \log (\frac{\#E}{\#E_w})$, where $\#E$ is the total number of log events and $\#E_w$ is the number of log events that contain $w$.
Indeed, there are some more advanced methods than GloVe and TF-IDF respectively, but our work does not aim to evaluate all of them.
If these simple methods can work well for improving the traditional PCA technique, we believe that more advanced ones can be helpful to further improve the effectiveness, which can be regarded as our future work.
Also, in the future, we can investigate the effectiveness of other log representation methods (such as synonym-antonym-based representation) in improving traditional techniques.

\subsubsection{Anomaly Detection Model Building}
\label{sec:build}
An anomaly detection model is constructed based on log sequences.
Hence, after obtaining log event vectors,
it requires to further represent a log sequence as a vector.
Existing log sequence representation methods either simply aggregate log event vectors via addition or incorporate advanced DL models to represent the log sequences.
Since the former is more lightweight, \tech{} adopts it for log sequence representation in order to keep the advantages of its high efficiency and weak data dependency.
In the future, more log sequence representation methods can be explored.


Based on the log sequence vectors, \tech{} adopts the PCA algorithm to build an anomaly detection model.
The core of PCA in anomaly detection is to build a \textbf{normal space} (denoted as $S_n$, and an \textbf{abnormal space} (denoted as $S_a$) as shown in the PCA step in Figure~\ref{fig:overview}.
Specifically, it projects each log sequence vector (denoted as $v_i$) into $y_n^{i}$ by $y_n^i=PP^Tv_i$, where $P=[e_1, e_2, \dots e_k]$ and $e_1, e_2, \dots e_k$ refer to the eigenvectors with the $k$ largest eigenvalues.
That also indicates that there are $k$ principal components.
We call $y_n^{i}$ the projection of $v_i$ to the normal space ($S_n$).
Similarly, it also projects $v_i$ to the abnormal space via $y_a^i = (I-PP^T)v_i$, where $I$ refers to the standard identity matrix.
Since the components corresponding to the abnormal space retain the minimal variance of the original data, \tech{} detects anomalies through outlier detection in the abnormal space ($S_a$).
Following the existing work~\cite{PCA}, \tech{} adopts Square Prediction Error (SPE)~\cite{dunia1997multi} to determine whether an incoming log sequence (denoted as $v_j$) is an outlier or not in the abnormal space by comparing with a given threshold $\theta$ (which is a hyper-parameter in \tech{}).
Specifically, the SPE value for $v_j$ is calculated by $||y_a^j||^2$.
If the SPE value is larger than $\theta$, $v_j$ is predicted as an anomalous log sequence.

\subsection{Datasets}
\label{sec:dataset}
In our study, we evaluated the studied log-based anomaly detection techniques on five datasets, including three public datasets (i.e., \HDFS{}~\cite{HDFS}, \BGL{}~\cite{BGL}, \Spirit{}~\cite{Spirit}) that have been widely-used in the existing work~\cite{DeepLog,LogAnomaly,PLELog,NeuralLog}, as well as two industrial datasets
from two large-scale distributed online systems in two different organizations, i.e., the network center of our university and one influential motor corporation throughout the world.
We hide the company name and system names due to the confidential policy.
For ease of presentation, we call the two industrial datasets \NC{} and \ZQY{}, respectively.

Table~\ref{tab:dataset_statistics} shows the basic information of these datasets, including the number of log messages, the used method of log data grouping (i.e., log sequence construction by grouping log messages), the average length of a log sequence, the number of log sequences in the training/validation/test sets, the ratio of anomalous log sequences in the training/validation/test sets, the total number of log events, and the number of unseen log events in the test set.
In the following, we will introduce each dataset in detail.

\begin{table}[t]
  \centering
  \small
  \caption{Basic information of datasets. The window size of the Fixed Window strategy refers to the number of log messages within each window, while the window size of the Fixed Time Window strategy refers to the time period within which all log messages are grouped as a log sequence.}
\resizebox{\textwidth}{!}{%
\begin{tabular}{@{}crrrrrrrrrrr@{}}
\toprule
\multicolumn{1}{c}{\multirow{2}{*}{Datasets}} & \multicolumn{1}{c}{\multirow{2}{*}{\begin{tabular}[c]{@{}c@{}}\# Log \\ Messages\end{tabular}}} & \multicolumn{1}{c}{\multirow{2}{*}{\begin{tabular}[c]{@{}c@{}}Grouping \\ Strategy\end{tabular}}}       & \multicolumn{1}{c}{\multirow{2}{*}{\begin{tabular}[c]{@{}c@{}}avg. \\ Length\end{tabular}}} & \multicolumn{3}{c}{\# Sequences}                                                                          & \multicolumn{3}{c}{\% Anomaly}                                                                            & \multicolumn{1}{c}{\multirow{2}{*}{\# Events}} & \multicolumn{1}{c}{\multirow{2}{*}{\# Unseen}} \\
\multicolumn{1}{c}{}                          & \multicolumn{1}{c}{}                                                                            & \multicolumn{1}{c}{}                                                                                    & \multicolumn{1}{c}{}                                                                        & \multicolumn{1}{c}{\textit{train}} & \multicolumn{1}{c}{\textit{val}} & \multicolumn{1}{c}{\textit{test}} & \multicolumn{1}{c}{\textit{train}} & \multicolumn{1}{c}{\textit{val}} & \multicolumn{1}{c}{\textit{test}} & \multicolumn{1}{c}{}                           & \multicolumn{1}{c}{}                           \\ \midrule
\HDFS{}                                          & 11,175,629                                                                                      & \begin{tabular}[c]{@{}r@{}}Session Window\\ (block\_id)\end{tabular}                                    & 19                                                                                          & 345,036                            & 57,506                           & 172,518                           & 3.27                               & 3.20                             & 2.15                              & 46                                             & 1                                              \\\midrule
\BGL{}                                           & 4,747,963                                                                                       & \begin{tabular}[c]{@{}r@{}}Session Window\\ (node\_id)\\ + Fixed Window\\ (size = 120)\end{tabular} & 55                                                                                          & 51,346                             & 8,557                            & 25,673                            & 52.96                              & 53.21                            & 17.75                             & 431                                            & 40                                             \\\midrule
\Spirit{}                                        & 12,000,000                                                                                      & \begin{tabular}[c]{@{}r@{}}Fixed Window\\ (size = 120)\end{tabular}                                     & 120                                                                                         & 60,000                             & 10,000                           & 30,000                            & 25.74                              & 25.39                            & 29.98                             & 1414                                           & 53                                             \\\midrule
\NC{}                                            & 334,567                                                                                         & \begin{tabular}[c]{@{}r@{}}Fixed Time Window\\ (size = 5min)\end{tabular}                                    & 13                                                                                          & 15,441                             & 2,573                            & 7,721                             & 21.43                              & 6.86                             & 2.19                              & 23                                             & 12                                             \\\midrule
\ZQY{}                                            & 6,148,033                                                                                       & \begin{tabular}[c]{@{}r@{}}Fixed Time Window\\ (size = 5min)\end{tabular}                                    & 210                                                                                         & 17,550                             & 2,925                            & 8,775                             & -                                  & -                                & -                                 & 245                                            & 31                                             \\ \bottomrule
\end{tabular}%
}
  \label{tab:dataset_statistics}
\end{table}%


\textbf{\HDFS{}} (Hadoop Distributed File System) was collected during the running of Hadoop-based MapReduce jobs on more than 2,000 Amazon's EC2 nodes for 38.7 hours.
It contains 11,175,629 log messages, which form 575,062 log sequences according to the \textit{block\_id} of each log message.
Among them, 2.9\% of log sequences are anomalous, which were manually labeled by Hadoop domain experts. 

\textbf{\BGL{}} (Blue Gene/L supercomputer) was produced by the Blue Gene/L supercomputer, which consists of 128K processors and was deployed at the Lawrence Livermore National Labs with a time span of 7 months.
In total, it contains 4,747,963 log messages, among which 348,460 are anomalous. 
Following the existing work~\cite{LogAnomaly,DeepLog,PLELog}, 
we combined the fixed window strategy and the session window strategy, which apply the fixed window strategy with the window size of 120 (this number refers to the number of log messages) to each set of log data with the same \textit{node\_id}.
As shown in the Table~\ref{tab:dataset_statistics}, the average length of a log sequence in this dataset is 55, which is smaller than the window size -- 120.
This is because the number of log messages with a certain \textit{node\_id} can be smaller than 120 and our fixed window strategy is applied to the log messages with the same \textit{node\_id}.
With this strategy, a log sequence is regarded as an anomalous one if it contains at least one anomalous log message.

\textbf{\Spirit{}} was produced by the Spirit supercomputer at Sandia National Labs~\cite{Spirit} with a time span of 2.5 years.
There are 272,298,969 log messages in total, including 181,642,697 anomalous messages.
Similar to the existing work~\cite{NeuralLog}, we sampled the first 12 million log messages (including 360,000 anomalous messages) in the whole dataset as the studied dataset in our work since using the whole dataset requires unaffordable computing and storage resources~\cite{NeuralLog}.
We also used the strategy of fixed windows with the window size of 120 to extract log sequences from log messages.

\textbf{\NC{}} was produced by a real-world large-scale online service system in the Network Center of our university. 
It contains several kinds of anomalies, such as system failures, network issues, and security breach.
In total, there are 334,567 log messages, among which 15,394 are anomalous.
The annotation process is semi-automatic, where system monitors first report alerts according to the pre-defined rules and operators then manually check them to filter out false alarms.
Based on the suggestions from its operators, we adopted the fixed time window strategy with the window size of 5 minutes to extract log sequences following the practical scenario of it.
That is, the log messages within a 5-minutes period are grouped as a log sequence.

\textbf{\ZQY{}} was produced by a real-world large-scale online service system in one influential motor corporation throughout the world, which focuses on developing, simulating, and testing various autonomous driving algorithms. 
There are 6,148,033 log messages in \ZQY{}, among which 4,833 are labeled as anomalous ones by domain experts following the similar annotation process as \NC{}.  
Similarly, we used the strategy of fixed time windows with the window size of 5 minutes to extract log sequences as well, confirmed by the operators of this system. 
Due to the confidential policy of the corporation, we have to omit the anomaly ratio of this dataset.

For each dataset, we split it into a training set, a validation set, and a test set in chronological order of log sequences with the ratio of 6:1:3.
Here, we did not adopt the widely-used splitting method in the existing work~\cite{DeepLog,LogRobust}, 
which shuffles all the log sequences before splitting.
This is because it could lead to data leakage~\cite{DBLP:conf/sigsoft/TuZZZ18} and thus evade the problem of unseen log events.
Our splitting method can ensure all the log sequences in the training set are produced before the log sequences in the test set, which is much closer to the practical scenario and indeed comes across the problem of unseen log events (especially for \BGL{} and \Spirit{} due to their longer time spans for log data).


\subsection{Metrics}
\label{sec:metric}
In our study, we measured both effectiveness and efficiency of each studied log-based anomaly detection technique.

Since log-based anomaly detection is actually a binary classification problem (i.e., distinguishing anomalies and normal cases), we adopted \textit{Precision} (\textit{Prec.}), \textit{Recall}, and \textit{F1-score} ($F_{1}$) as the metrics to measure the effectiveness of each technique following the existing work~\cite{DeepLog,LogRobust,PLELog,LogAnomaly,DBLP:conf/issre/HeZHL16}.
\textit{Precision} is computed by $\frac{TP}{TP+FP}$ while \textit{Recall} is computed by $\frac{TP}{TP+FN}$, where $\textit{TP}$, $\textit{FP}$, and $\textit{FN}$ refer to 
the number of true positives (an \textit{anomalous} log sequence is indeed predicted to be \textit{anomalous}), false positives (a \textit{normal} log sequence is predicted to be \textit{anomalous}), and false negatives (an anomalous log sequence is predicted to be \textit{normal}), respectively.
\textit{F1-score} considers both \textit{Precision} and \textit{Recall}, which is computed by $\frac{2 \cdot (\textit{Precision} \cdot \textit{Recall})}{\textit{Precision}+\textit{Recall}}$.

Furthermore, following the existing work~\cite{fse21nengwen,PLELog,menzies_easy_over_hard}, we adopted the time spent on building an anomaly detection model (called training time) and the time spent on online anomaly detection (called prediction time) as the efficiency metrics in our study. 
Although building an anomaly detection model is an offline task, the time spent on it is still important since the anomaly detection model has to be frequently updated due to the frequent software evolution.
Please note that training time does not include the time spent on hyper-parameter tuning through grid search, but refers to the time spent on build the anomaly detection model based on the best hyper-parameters after tuning.
It is reasonable to exclude the time spent on hyper-parameter tuning through grid search since each technique employs grid search for hyper-parameter tuning.
Prediction time refers to the average time spent on predicting an incoming log sequence across the entire test set for each dataset. The time spent on predicting incoming log sequences for online anomaly detection is also quite important since an anomaly detection model is required to monitor the software status in \textit{real time}.

It is also important to measure the average training time for one epoch (called one-epoch training time) for DL-based techniques since their efficiency could be improved after the initial training by training for a few epochs or incremental updates in practice.
Moreover, based on one-epoch training time, we can also conclude the efficiency superiority of \tech{} over DL-based techniques for the process of grid search.



\subsection{Implementations and Configurations}
\label{sec:imple}
In total, we studied seven log-based anomaly detection techniques.
For PLELog, we directly used the released implementation by the existing work~\cite{PLELog}.
For PCA and LogCluster, we adopted the implementations provided by the existing study~\cite{DBLP:conf/issre/HeZHL16}.
Regarding DeepLog~\cite{DeepLog}, LogAnomaly~\cite{LogAnomaly}, and LogRobust~\cite{LogRobust}, their implementations are not available and thus we re-implemented them according to the descriptions in their papers.
To ensure that our re-implementations are correct and our usage for the released implementations are correct, we conducted a small experiment on HDFS to reproduce the result of each technique according to their original experimental settings.
Indeed, for each technique, the difference between our result and the result provided by the original work is minor, i.e., the average F1-score difference is just 0.013 across all the six techniques.
Regarding \tech{}, we implemented it in Python 3.8 based on scikit-learn 0.24~\cite{sklearn}.

Following the practice in deep learning~\cite{ripley2007pattern,DBLP:conf/issre/WangCYZZQKLRGXD21},
\textit{we determined the hyper-parameter settings of each studied technique through extensive grid search on the corresponding validation set}, in order to make each studied technique achieve the best effectiveness (i.e., F1-score) on each dataset.
Please note that the grid search process is not part of \tech{} or any other studied techniques, but it is common practice in deep learning~\cite{DBLP:conf/icse/WangY0ZDZ21,DBLP:conf/kbse/ZhangWJYC22}. 
Hence, we employed it for all the studied techniques on each validation set for sufficient and fair comparison.
In particular, we developed a toolbox that integrates the implementations of all the seven studied techniques in order to promote future research and practical use.
In this toolbox, each technique can be invoked by simply specifying the corresponding configuration.
Our toolbox and all the experimental data (also including the parameter settings of each technique used in our study after grid search) can be found at our project homepage: \textbf{\url{https://github.com/YangLin-George/SemPCA}}.
We conducted all the experiments on a server with Ubuntu 18.04 LTS, Intel Xeon Gold 6240C CPU, 128GB RAM, and a NVIDIA RTX3090.





\subsection{Experimental Setup}
\label{sec:setup}
RQ1 aims to investigate whether optimizing the traditional PCA technique through lightweight adaptation can achieve comparable effectiveness with the widely-studied techniques (especially the state-of-the-art DL-based techniques).
To answer RQ1, we applied each technique to each dataset, and then measured \textit{Precision}, \textit{Recall}, and \textit{F1-score} to compare these techniques in terms of effectiveness.



Then, we investigated how these studied techniques suffer from the two above-mentioned limitations.
Specifically, RQ2 aims to investigate whether the optimized version of the traditional PCA technique can perform more stably under insufficient training data than the existing techniques.
To answer RQ2, for each dataset, we constructed a series of training sets by randomly sampling 1\%, 2\%, 5\%, 10\%, and 20\% of log sequences from the whole training data respectively, which aims to simulate the training sets with different degrees of quality by changing the amounts of training data.
Here, we constructed the training sets with much less training data, which aims to simulate the practical scenario suffering from the data hungry problem~\cite{menzies_easy_over_hard}.
Specifically, collecting sufficient training data (especially labeled data) is challenging and costly in practice, and thus the small amount of data is a common factor affecting the quality of training data.
If a technique can perform well with various small amounts of training data, it means that this is a stable technique without heavy dependency on training data.
Besides, for each studied amount, we repeated the sampling process 10 times for constructing 10 different training sets with the same amount of training data.
On the one hand, it can reduce the threat from randomness;
On the other hand, the training sets with the same amount of log data could also have different degrees of quality (e.g., different degrees of data imbalance), and thus such a way can help investigate the stability of each technique when controlling for the amount of training data.

Please note that for each dataset, all the log sequences in these constructed training sets are sampled from the whole training set and meanwhile we used the same test set 
for fair evaluation.
With these constructed training sets, we applied each technique to each of them and then measured its effectiveness on the test set of the corresponding dataset.

Subsequently, RQ3 aims to investigate the efficiency of these studied techniques.
To answer RQ3, we recorded the training time and prediction time for each technique on each dataset.

\section{Results and Analysis}
\label{sec:results}

\begin{table*}[t!]
    \centering
    \setlength\tabcolsep{3pt} 
    \small
    \caption{Effectiveness comparison among the studied techniques}
    \label{tab:rq1_main}
    \resizebox{\textwidth}{!}{%
    \begin{tabular}{l|ccc|ccc|ccc|ccc|ccc}
    \toprule
    \multirow{2}{*}{Technique} & \multicolumn{3}{c|}{\uline{\HDFS{}}}                                              & \multicolumn{3}{c|}{\uline{\BGL{}}}                                 &\multicolumn{3}{c|}{\uline{\Spirit{}}}              & \multicolumn{3}{c|}{\uline{\ZQY{}}}                                                & \multicolumn{3}{c}{\uline{\NC{}}}                                                                                             \\
                              & \multicolumn{1}{c}{\textit{Prec.}} & \multicolumn{1}{c}{\textit{Recall}} & \multicolumn{1}{c|}{$F_{1}$} & \multicolumn{1}{c}{\textit{Prec.}} & \multicolumn{1}{c}{\textit{\textit{Recall}}} & \multicolumn{1}{c|}{$F_{1}$} & \multicolumn{1}{c}{\textit{Prec.}} & \multicolumn{1}{c}{\textit{\textit{Recall}}} & \multicolumn{1}{c|}{$F_{1}$} & \multicolumn{1}{c}{\textit{Prec.}} & \multicolumn{1}{c}{\textit{Recall}} & \multicolumn{1}{c|}{$F_{1}$} & \multicolumn{1}{c}{\textit{Prec.}} & \multicolumn{1}{c}{\textit{\textit{Recall}}} & \multicolumn{1}{c}{$F_{1}$} \\ \midrule
PCA                       & 0.996          & 0.815          & 0.897          & 0.565          & 1.000          & 0.722          & 0.924          & 0.945          & 0.934          & 0.931          & 0.979          & 0.954          & 0.966          & 0.821          & 0.887          \\
LogCluster                & 0.997          & 0.900          & 0.946          & 0.988          & 0.623          & 0.766          & \textbf{0.994} & 0.942          & 0.968          & \textbf{0.991} & 0.879          & 0.932          & 0.999          & 0.671          & 0.802          \\ \midrule
DeepLog                   & 0.864          & 0.958          & 0.909          & 0.162          & 0.868          & 0.273          & 0.889          & \textbf{0.995} & 0.939          & 0.946          & 0.999          & 0.972          & 0.996          & 0.994          & 0.995          \\
LogAnomaly                & 0.933          & 0.992          & 0.962          & 0.151          & 0.791          & 0.253          & 0.883          & 0.975          & 0.927          & 0.945          & 0.999          & 0.971          & 0.979          & 0.997          & 0.988          \\
PLELog                    & 0.989          & 0.957          & 0.973          & 0.978          & 0.998          & 0.988          & 0.963          & 0.980          & \textbf{0.971} & 0.985          & 0.994          & 0.984          & \textbf{1.000} & \textbf{0.999} & \textbf{0.999} \\
LogRobust                 & \textbf{0.999} & 0.997          & \textbf{0.998} & \textbf{0.999} & 0.999          & \textbf{0.999} & 0.994          & 0.892          & 0.938          & 0.988          & 0.982          & \textbf{0.985} & 0.999          & 0.994          & 0.995          \\ \midrule
\ins{\tech{}}\del{PCA++}                    & 0.963          & \textbf{1.000} & 0.981          & 0.897          & \textbf{1.000} & 0.946          & 0.981          & 0.920          & 0.950          & 0.932          & \textbf{1.000} & 0.965          & 0.997          & 0.912          & 0.953          \\ \bottomrule
\end{tabular}
    }
    
\end{table*}


\subsection{RQ1: Effectiveness Comparison}
\label{sec:RQ1}
Table~\ref{tab:rq1_main} shows the comparison results of the seven studied techniques in terms of anomaly detection effectiveness, where the best effectiveness on each dataset is marked as bold.
By comparing \tech{} with PCA, we found that the former performs significantly better than the latter, e.g., the average improvement in terms of F1-score is 0.101 across all the datasets.
\ins{\textbf{The results demonstrate the significant contribution of the semantic-based log representation method in \tech{}.}}
Moreover, \tech{} also outperforms another traditional technique (i.e., LogCluster) on all the datasets (except \Spirit{}).
For example, the improvements in terms of F1-score on \HDFS{}, \BGL{}, \ZQY{}, and \NC{} range from 3.541\% to 23.822\%.

Regarding the studied DL-based techniques, the supervised technique (i.e., LogRobust) and the state-of-the-art semi-supervised technique (i.e., PLELog) perform better than the other two (i.e., DeepLog and LogAnomaly).
This is as expected since both LogRobust and PLELog carefully handle the problem of unseen log events by designing the semantic-based log representation methods and meanwhile LogRobust incorporates much more information (i.e., the labels of both normal and anomalous log sequences).
Surprisingly, the optimized version of the \textit{unsupervised} PCA technique (i.e., \tech{}) achieves comparable effectiveness with both LogRobust and PLELog, even though \tech{} does not require any label information and relies on a traditional algorithm (i.e., PCA) for building an anomaly detection model.
The F1-score differences between \tech{} and PLELog only range from 0.021 to 0.046 for \BGL{}, \Spirit{}, \ZQY{}, and \NC{}, and those between \tech{} and LogRobust only range from 0.017 to 0.053 for \HDFS{}, \BGL{}, \ZQY{}, and \NC{}.
Even \tech{} performs slightly better than LogRobust on \Spirit{} and PLELog on \HDFS{} in terms of F1-score.
The reason why LogRobust does not perform very well on \Spirit{} may be that LogRobust is supervised, i.e., learning patterns from both normal and abnormal log sequences, but the imbalance problem on \Spirit{} is very significant.
The results demonstrate the great potential of \tech{} in log-based anomaly detection, indicating that properly optimizing traditional techniques is also a promising direction in this area.

From Table~\ref{tab:rq1_main}, we also found that on \BGL{}, both DeepLog and LogAnomaly perform significantly worse than all the other techniques (even the traditional ones, i.e., PCA and LogCluster).
This is because the \BGL{} dataset suffers from the most serious unseen-log-event problem among all the studied datasets. 
Both DeepLog and LogAnomaly predict several possible log events that can follow the subsequence of log events in a log sequence.
In this way, when coming across an unseen log event that is dissimilar to the predicted ones according to the log event vectors, the log sequence is directly identified as an anomaly. 
Although LogAnomaly also extracts semantics of log events through synonyms-antonyms-based log representation, the way of predicting an anomaly makes it hard to handle the problem of unseen log events.
On the other hand, although PCA and LogCluster do not handle this problem, they directly ignore unseen log events and count the occurrence of seen log events for anomaly detection, which is less aggressive to treat the log sequences with unseen log events than DeepLog and LogAnomaly.
Indeed, all these techniques without well relieving the problem of unseen log events perform much worse than those incorporating semantic-based log representation on \BGL{}, which further confirms the necessity of relieving this problem in log-based anomaly detection.

\finding{\tech{} significantly outperforms PCA by incorporating lightweight semantic-based log representation, and achieves comparable effectiveness with the supervised DL-based technique (i.e., LogRobust) and the semi-supervised DL-based technique (i.e., PLELog).}

\begin{table*}[t!]
    \centering
    \setlength\tabcolsep{3pt} 
    \small
    \caption{Average and standard deviation of F1-score of each studied technique under different amounts of training data.}
    \label{tab:rq2_statistics}
    \resizebox{\textwidth}{!}{
    \begin{tabular}{@{}l|l|cc|cc|cc|cc|cc|cc|cc@{}}
\toprule
                          &                            & \multicolumn{2}{c|}{{\ul PCA}}                            & \multicolumn{2}{c|}{{\ul LogCluster}}                     & \multicolumn{2}{c|}{{\ul DeepLog}}                        & \multicolumn{2}{c|}{{\ul LogAnomaly}}                         & \multicolumn{2}{c|}{{\ul PLELog}}                         & \multicolumn{2}{c|}{{\ul LogRobust}}                      & \multicolumn{2}{c}{{\ul \tech{}}}                           \\
\multirow{-2}{*}{Dataset} & \multirow{-2}{*}{Ratio}    & avg.                          & $\sigma$                     & avg.                          & $\sigma$                     & avg.                          & $\sigma$                     & avg.                          & $\sigma$                         & avg.                          & $\sigma$                     & avg.                          & $\sigma$                     & avg.                           & $\sigma$                     \\ \midrule
                          & 1                          & 0.872                         & 0.022                     & 0.946                         & 0.001                     & 0.853                         & 0.043                     & 0.932                         & 0.015                         & 0.767                         & 0.363                     & 0.924                         & 0.036                     & 0.980                         & 0.001                     \\
                          & 2                          & 0.881                         & 0.023                     & 0.947                         & 0.001                     & 0.822                         & 0.074                     & 0.944                         & 0.014                         & 0.979                         & 0.009                     & 0.963                         & 0.019                     & 0.978                         & 0.007                     \\
                          & 5                          & 0.884                         & 0.019                     & 0.947                         & 0.000                     & 0.784                         & 0.028                     & 0.795                         & 0.057                         & 0.785                         & 0.083                     & 0.981                         & 0.011                     & 0.981                         & 0.000                     \\
                          & 10                         & 0.888                         & 0.017                     & 0.947                         & 0.001                     & 0.875                         & 0.060                     & 0.794                         & 0.032                         & 0.727                         & 0.046                     & 0.989                         & 0.006                     & 0.976                         & 0.009                     \\
                         \multirow{-5}{*}{\HDFS{}}   & 20                         & 0.888                         & 0.016                     & 0.947                         & 0.001                     & 0.893                         & 0.045                     & 0.843                         & 0.051                         & 0.638                         & 0.201                     & 0.997                         & 0.001                     & 0.981                         & 0.000                     \\ \midrule
                          & 1                          & 0.709                         & 0.007                     & 0.670                         & 0.029                     & 0.267                         & 0.017                     & 0.256                         & 0.010                         & 0.755                         & 0.061                     & 0.718                         & 0.163                     & 0.908                         & 0.038                     \\
                          & 2                          & 0.714                         & 0.013                     & 0.695                         & 0.029                     & 0.249                         & 0.013                     & 0.237                         & 0.015                         & 0.667                         & 0.186                     & 0.911                         & 0.038                     & 0.928                         & 0.033                     \\
                          & 5                          & 0.714                         & 0.014                     & 0.741                         & 0.017                     & 0.244                         & 0.015                     & 0.227                         & 0.001                         & 0.870                         & 0.076                     & 0.914                         & 0.105                     & 0.941                         & 0.004                     \\
                          & 10                         & 0.719                         & 0.010                     & 0.757                         & 0.008                     & 0.241                         & 0.017                     & 0.227                         & 0.000                         & 0.961                         & 0.030                     & 0.963                         & 0.027                     & 0.937                         & 0.006                     \\
                      \multirow{-5}{*}{\BGL{}}    & 20                         & 0.717                         & 0.010                     & 0.762                         & 0.005                     & 0.236                         & 0.016                     & 0.227                         & 0.000                         & 0.961                         & 0.028                     & 0.972                         & 0.027                     & 0.945                         & 0.005                     \\ \midrule
                          & 1                          & 0.073                         & 0.004                     & 0.904                         & 0.035                     & 0.732                         & 0.028                     & 0.918                         & 0.010                         & 0.809                         & 0.101                     & 0.227                         & 0.321                     & 0.916                         & 0.036                     \\
                          & 2                          & 0.205                         & 0.266                     & 0.924                         & 0.015                     & 0.708                         & 0.067                     & 0.938                         & 0.012                         & 0.700                         & 0.093                     & 0.313                         & 0.330                     & 0.919                         & 0.032                     \\
                          & 5                          & 0.325                         & 0.311                     & 0.947                         & 0.019                     & 0.663                         & 0.084                     & 0.957                         & 0.018                         & 0.836                         & 0.102                     & 0.294                         & 0.311                    & 0.923                         & 0.030                     \\
                          & 10                         & 0.595                         & 0.359                     & 0.955                         & 0.016                     & 0.614                         & 0.087                     & 0.931                         & 0.082                         & 0.857                         & 0.083                     & 0.363                         & 0.338                     & 0.933                         & 0.002                     \\
                   \multirow{-5}{*}{\Spirit{}}       & 20                         & 0.894                         & 0.080                     & 0.963                         & 0.011                     & 0.588                         & 0.060                     & 0.951                         & 0.033                         & 0.853                         & 0.053                     & 0.606                         & 0.296                     & 0.924                         & 0.030                     \\ \midrule
                          & 1                          & 0.943                         & 0.011                     & 0.947                         & 0.014                     & 0.989                         & 0.001                     & 0.856                         & 0.007                         & 0.963                         & 0.040                     & 0.788                         & 0.306                     & 0.964                         & 0.001                     \\
                          & 2                          & 0.941                         & 0.016                     & 0.942                         & 0.011                     & 0.989                         & 0.001                     & 0.867                         & 0.005                         & 0.979                         & 0.038                     & 0.947                         & 0.033                     & 0.963                         & 0.001                     \\
                          & 5                          & 0.944                         & 0.018                     & 0.937                         & 0.011                     & 0.989                         & 0.001                     & 0.874                         & 0.002                         & 0.922                         & 0.039                     & 0.976                         & 0.036                     & 0.964                         & 0.001                     \\
                          & 10                         & 0.952                         & 0.014                     & 0.936                         & 0.005                     & 0.989                         & 0.001                     & 0.875                         & 0.001                         & 0.890                         & 0.047                     & 0.952                         & 0.057                     & 0.964                         & 0.001                     \\
                    \multirow{-5}{*}{\ZQY{}}        & 20                         & 0.955                         & 0.010                     & 0.932                         & 0.004                     & 0.989                         & 0.001                     & 0.876                         & 0.001                         & 0.912                         & 0.029                     & 0.911                         & 0.052                     & 0.964                         & 0.000                     \\  \midrule
                          & 1                          & 0.882                         & 0.000                     & 0.789                         & 0.000                     & 0.970                         & 0.008                     & 0.978                         & 0.003                         & 0.963                         & 0.042                     & 0.979                         & 0.012                     & 0.948                         & 0.033                     \\
                          & 2                          & 0.882                         & 0.000                     & 0.789                         & 0.000                     & 0.972                         & 0.009                     & 0.980                         & 0.004                         & 0.979                         & 0.040                     & 0.990                         & 0.006                     & 0.935                         & 0.030                     \\
                          & 5                          & 0.883                         & 0.002                     & 0.789                         & 0.000                     & 0.968                         & 0.006                     & 0.983                         & 0.002                         & 0.922                         & 0.041                     & 0.994                         & 0.002                     & 0.980                         & 0.004                     \\
                          & 10                         & 0.884                         & 0.003                     & 0.789                         & 0.000                     & 0.980                         & 0.009                     & 0.989                         & 0.001                         & 0.890                         & 0.049                     & 0.997                         & 0.001                     & 0.980                         & 0.004                     \\
                        \multirow{-5}{*}{\NC{}}      & 20                         & 0.885                         & 0.003                     & 0.789                         & 0.000                     & 0.978                         & 0.009                     & 0.989                         & 0.001                         & 0.912                         & 0.031                     & 0.998                         & 0.001                     & 0.980                         & 0.004                      \\ \bottomrule
\end{tabular}
    }
\end{table*}
\subsection{RQ2: Stability Comparison}
\label{sec:rq2}

Table~\ref{tab:rq2_statistics} shows the effectiveness of each studied technique under different amounts of training data.
For each technique on each dataset under each studied amount of training data, we calculated the average F1-score (avg. denoted in the table) across 10 training sets with the same amount of training data, and calculated the corresponding standard deviation ($\sigma$ denoted in the table).
Here, we mainly studied small amounts of training data in order to simulate the common data-hungry problem in the practical use of deep learning~\cite{menzies_easy_over_hard}.


From Table~\ref{tab:rq2_statistics}, in general, the studied DL-based techniques perform much worse under small amounts of training data than using the whole training data (shown in Table~\ref{tab:rq1_main}) in terms of average F1-score.
This is as expected since DL-based techniques tend to rely on a huge amount of training data to learn the parameters in complex neural networks~\cite{access10review,csur18survey}.
In contrast, the traditional techniques (i.e., \tech{}, PCA, LogCluster) perform stably regardless of the studied amounts of training data.
In particular, when using 1\% of training data, the optimized version of the unsupervised PCA technique (i.e., \tech{}) performs the best among these studied techniques in most cases.
For example, the unsupervised \tech{} performs much better than the supervised LogRobust in four (out of five) datasets under 1\% of training data, i.e., the improvements of \tech{} over LogRobust in terms of average F1-score are 6.061\% on \HDFS{}, 26.462\% on \BGL{}, 136.082\% \Spirit{}, and 22.335\% on \ZQY{}.
The results demonstrate that \tech{} is more stable and less dependent on the amount of training data than the advanced DL-based techniques.

Please note that even though 1\% is a very small ratio, 1\% of training data does not mean the very small amount of training data.
For example, for \HDFS{}, 1\% of training data contains 57,506 log sequences.
In practice, labeling such a large number of training data is still very expensive, indicating that the data hungry problem could be more serious in practice and thus the effectiveness of these DL-based techniques could be affected more deeply.



From the results of standard deviation in Table~\ref{tab:rq2_statistics}, \tech{} (as well as the other two traditional techniques) perform more stably than the studied DL-based techniques.
For example, on \Spirit{}, the standard deviation of \tech{} ranges from 0.002 to 0.036 while those of the supervised LogRobust and the state-of-the-art semi-supervised PLELog range from 0.280 to 0.335 and from 0.053 to 0.102 respectively, across all the studied amounts of training data.
The mean of standard deviation for \tech{} is 0.012 across all the cases, which is 88.000\%, 84.211\%, 57.143\%, and 33.333\% smaller than that of LogRobust, PLELog, DeepLog, and LogAnomaly respectively.
The results demonstrate that the optimized version of the traditional PCA technique (i.e., \tech{}) is much more stable than the studied DL-based techniques when controlling for the amount of training data, indicating that these studied DL-based techniques also more heavily depend on the quality of training data.
In practice, collecting and labeling sufficient \textit{high-quality} training data is quite challenging, which further hinders the practical use of these DL-based techniques.

In particular, LogRobust performs significantly worse and less stably than all the other techniques on \Spirit{} under small amounts of training data.
The main reason lies in that LogRobust requires to learn patterns from both normal and abnormal log sequences and meanwhile the distribution of the data in \Spirit{} is more diverse, and thus it requires much more training data for building an accurate anomaly detection model.
Indeed, when using the whole training data, the effectiveness of LogRobust is significantly improved.
However, all the other techniques either just learn patterns from normal log sequences or detect anomalies via outlier analysis, and thus the dependency on the training data amount is relatively small.
The results further confirm the challenge of applying the supervised log-based anomaly detection techniques to the practice.

\finding{The optimized version of the traditional PCA technique (i.e., \tech{}) is more stable and less dependent on the amount and quality of training data than the advanced DL-based techniques.
When using small amounts of training data, the former even performs better than the latter in terms of average F1-score.}



\subsection{RQ3: Efficiency Comparison}
\label{sec:rq3}

\begin{table}[t]
\small
    \caption{\ins{Efficiency comparison among the studied techniques.}}
    \label{tab:efficiency}
\centering
\resizebox{\textwidth}{!}{%
\begin{tabular}{@{}cc|r|r|r|r|r|r|r@{}}
\toprule
\multicolumn{2}{c|}{Dataset}    & \multicolumn{1}{c|}{PCA} & \multicolumn{1}{c|}{LogCluster} & \multicolumn{1}{c|}{DeepLog} & \multicolumn{1}{c|}{LogAnomaly} & \multicolumn{1}{c|}{LogRobust} & \multicolumn{1}{c|}{PLELog} & \multicolumn{1}{c}{SemPCA} \\ \midrule
\multirow{3}{*}{\HDFS{}}   & \textit{train}   & 0.103                    & 10.195                          & 1,782.453                    & 72,933.082                      & 3,252.943                       & 3,782.365                   & 1.053                      \\
                        & \textit{one-epoch}    & -                        & -                               & 44.562                       & 1,823.327                        & 162.602                        & 189.118                     & -                          \\
                        & \textit{pred.}    & 0.002                    & 0.046                           & 0.042                        & 0.048                           & 0.200                          & 0.197                       & 0.003                      \\ \midrule
\multirow{3}{*}{\BGL{}}    & \textit{train}   & 0.118                    & 6.859                           & 645.082                      & 36,594.352           & 1,016.560                      & 998.542                     & 0.101                      \\
                        & \textit{one-epoch}    & -                        & -                               & 16.127                       & 3,148.195                       & 50.828                         & 49.925                      & -                          \\
                        & \textit{pred.}    & 0.026                    & 0.437                           & 0.384                        & 0.282                           & 0.367                          & 0.268                       & 0.028                      \\ \midrule
\multirow{3}{*}{\Spirit{}} & \textit{train}   & 0.983                    & 48.936                          & 2,566.723                    & \textgreater 24 hours           & 2,361.429                      & 2,855.766                   & 0.164                      \\
                        & \textit{one-epoch}    & -                        & -                               & 64.168                       & 36,821.91                       & 118.071                        & 142.788                     & -                          \\
                        & \textit{pred.}    & 0.303                    & 0.611                           & 19.326                       & 152.793                         & 1.704                          & 1.251                       & 0.016                      \\ \midrule
\multirow{3}{*}{\ZQY{}}     & \textit{train}   & 0.011                    & 0.075                           & 29.847                       & 1,862.281                       & 57.967                         & 56.747                      & 0.017                      \\
                        & \textit{one-epoch}    & -                        & -                               & 0.746                        & 46.557                          & 2.898                          & 2.837                       & -                          \\
                        & \textit{pred.}    & 0.014                    & 0.018                           & 1.131                        & 1.222                           & 0.152                          & 0.114                       & 0.003                      \\ \midrule
\multirow{3}{*}{\NC{}}     & \textit{train}   & 0.017                    & 2.955                           & 1,637.920                    & 33,975.866                      & 1,414.442                      & 922.881                     & 0.061                      \\
                        & \textit{one-epoch}    & -                        & -                               & 40.948                       & 849.375                         & 70.722                         & 46.144                      & -                          \\
                        & \textit{pred.}    & 0.068                    & 0.017                           & 0.798                        & 0.194                           & 0.871                          & 0.759                       & 0.015                      \\ \midrule
\multirow{3}{*}{Avg.}   & \textit{train}   & 0.246                    & 13.804                          & 1,332.405                    & 58,674.582                      & 1,620.668                      & 1,723.260                   & 0.279                      \\
                        & \textit{one-epoch}    & -                        & -                               & 33.310                       & 8,537.873                       & 81.024                         & 86.162                      & -                          \\
                        & \textit{pred.}    & 0.0826                   & 0.226                           & 4.336                        & 30.908                          & 0.659                          & 0.518                       & 0.013                      \\ \bottomrule
\end{tabular}%
}

\end{table}
Table~\ref{tab:efficiency} shows the efficiency comparison results among these studied techniques.
For each technique, we reported \ins{the training time (\textit{train}, in seconds), the prediction time (\textit{pred}, in milliseconds), the one-epoch training time (\textit{one-epoch}, in seconds), on each dataset.}\del{average training time and the average prediction time across all the datasets, respectively.}

From this table, the traditional techniques spend significantly shorter training time than the advanced DL-based techniques \del{on average}\ins{on each dataset}.
\del{For example, the average training time of \tech{} is only 0.226 second while that of DeepLog, LogAnomaly, PLELog, and LogRobust is 427.545, 13,363.516, 1,776.314, and 1,166.621 seconds respectively.}\ins{For example, the average training time of \tech{} is only 0.279 seconds while that of DeepLog, LogAnomaly, PLELog, and LogRobust are 1,332.405, 58,674.582, 1,723.260, and 1,620.668 seconds, respectively.
Among the four DL-based techniques, LogAnomaly was more than 10 times slower than the others, especially on the \Spirit{} dataset.
This is because LogAnomaly predicts the next log event based on the previous ones, causing that each log sequence can be transformed to multiple instances and thus the volume of training data can be significantly increased.
Moreover, the inclusion of two LSTMs (one incorporating quantity features and another incorporating semantic features) further aggravates the training cost.}
\ins{Moreover, the training time for \tech{} is stable for different datasets with different scales of training data (ranging from 0.017 seconds to 1.053 seconds), while that for DL-based techniques is affected significantly (ranging from 29.847 seconds to 2,566.723 seconds for DeepLog, ranging from 1,862.281 seconds to 148,004.419 seconds for LogAnomaly, ranging from 57.967 seconds to 3,252.943 seconds for LogRobust, and ranging from 56.747 seconds to 3,782.365 seconds for PLELog).
The results further demonstrate the superiority of \tech{} in terms of efficiency across different datasets.}

\ins{
Furthermore, we found that the time spent on the whole process of building an anomaly detection with \tech{} is still significantly shorter than the one-epoch training time for each DL-based technique.
For example, the whole training time for \tech{} is 0.279 seconds, while the one-epoch training time for LogRobust is 81.024 seconds, on average across all the datasets. 
That is, even though the efficiency of DL-based techniques could be improved after the initial training by training for a few epochs or incremental updates, \tech{} is still much more efficient than DL-based techniques.
Based on these results, we can also infer that the grid-search process for DL-based techniques can be more time-consuming than that for \tech{} (since an anomaly detection model has to be built and evaluated under each explored hyper-parameter setting in the grid search process), which further indicates the efficiency of \tech{}.
}

In terms of average prediction time, that of all the studied techniques is acceptable, and \tech{} and PCA spend shorter prediction time than the other techniques.


To sum up, \tech{} and PCA are much more efficient than the advanced DL-based techniques in terms of both training and prediction time.
Such high efficiency is helpful to update the anomaly detection model in time, tune the hyper-parameters sufficiently and efficiently, and monitor the system in real time.
The results demonstrate the practicability of \tech{}, especially when comprehensively considering the effectiveness, efficiency, and stability.

\finding{All the studied techniques have acceptable prediction time and that of \tech{} and PCA is shorter. However, the traditional techniques (i.e., \tech{}, PCA, and LogCluster) spend significantly shorter training time than the advanced DL-based techniques. This also confirms that incorporating lightweight semantic-based log representation into the PCA technique does not damage its advantage of high efficiency.}

\textbf{\underline{Summary}}: The unsupervised \tech{} (i.e., the optimized version of the traditional PCA technique through simple adaptation) achieves comparable effectiveness with the supervised and semi-supervised DL-based techniques while is more stable, less dependent on training data, and more efficient.
That demonstrates the great potential of \tech{} for more practical log-based anomaly detection, and confirms that such a traditional technique can still excel after simple but useful adaptation.

\vspace{1mm}
\textbf{\underline{Implication}}:
Please note that the point of our study is not to deprecate deep learning in the area of log-based anomaly detection, but evaluate the existing DL-based techniques from a novel perspective.
That is, we investigated whether the existing DL-based techniques have prominent advantage compared with the traditional techniques through simple but useful adaptation in terms of comprehensive metrics (i.e., effectiveness, efficiency, and stability).
Although our conclusions are relatively negative for the existing DL-based techniques in terms of the studied metrics, it is still exciting to see new DL-based techniques that can significantly outperform all the existing techniques (including the traditional techniques with some optimizations) in terms of effectiveness but cannot further damage the efficiency and stability under insufficient training data.
In the future, when evaluating newly-proposed DL-based techniques, optimizing the traditional techniques accordingly for comparison should be also noticed in order to confirm the contribution of advanced DL models to the significant effectiveness improvement.

\section{Discussion}
\label{sec:dis}
\subsection{Why does \tech{} Work?}
Based on our study, the traditional PCA algorithm can still excel by incorporating a lightweight semantic-based log representation in log-based anomaly detection.
The success of \tech{} is inspiring, and we further analyzed the insight behind its success.
The main reasons are threefold.

First, incorporating semantic-based log representation is helpful to relieve the problem of unseen log events.
The original PCA represents a log sequence as a event count vector by counting the occurrence of each log event (seen in training data) in the log sequence, which directly ignores the unseen log events in incoming log sequences and thus negatively affects its effectiveness.
However, the semantic-based log representation method in \tech{} can map the unseen log events in incoming log sequences to the log events seen in training data by measuring their semantic similarity.

Second, the semantic-based log representation method in \tech{} can facilitate the understanding of semantic differences between anomalous and normal log sequences, and thus \tech{} is effective to detect more obscure anomalies.
However, the event count vector method in the original PCA is just effective to detect the anomalies exhibited by the exceptional number of log events, which are usually obvious and could be also captured by the semantic differences.

Third, the number of anomalies tends to be much smaller than that of normal cases in practice~\cite{DeepLog,LogAnomaly,PLELog}, leading to the problem of imbalanced data.
Such imbalanced data could negatively affect the effectiveness of supervised DL models, which has been widely recognized in the area of deep learning~\cite{DBLP:journals/air/BejaniG21,DBLP:journals/corr/abs-2201-03299}, while it has little influence on the unsupervised PCA algorithm since PCA can effectively learn patterns from the majority class (i.e., normal log sequences) and then determine whether an incoming log sequence is an outlier with respect to the learned patterns.
Therefore, in our study, \tech{} performs more stably under a variety of training sets.

In addition, the adopted lightweight semantic-based log representation method in \tech{} does not consider the order information of log events, and thus the optimized version of the traditional PCA technique cannot detect the anomalies that are only reflected by log-event orders.
In fact, the existing study analyzed tens of thousands of anomalies related to logs in industry and found that (1) such a kind of anomalies are not common in practice;
(2) the anomalies involving abnormal log-event orders tend to also involve abnormal log-event counts or log-event types~\cite{fse21nengwen}.
Hence, many of such a kind of anomalies can be also detected by \tech{}.
In the future, we can further optimize traditional techniques by incorporate the order information of log events.

\subsection{Can Other Traditional Techniques Excel through Semantic-based Log Representation?}
\label{sec:other}
In the study, we took PCA as the representative to investigate whether traditional techniques through some optimizations can achieve comparable effectiveness with advanced DL-based techniques in log-based anomaly detection.
Indeed, we found that incorporating semantic-based log representation is helpful to largely improve the effectiveness of PCA almost without damaging its efficiency, since it can relieve the problem of unseen log events well.
Here, we further investigated whether the conclusion can be generalized to other traditional techniques. 
Specifically, we conducted \del{a small}\ins{an} experiment on the other two traditional techniques, i.e., LogCluster~\cite{LogCluster} and SVM~\cite{SVM}.
The former is semi-supervised \ins{that has been introduced in Section~\ref{sec:traditional}}, while the latter is supervised that first uses the event count vector method for log representation like PCA and LogCluster and then builds a classifier via the SVM algorithm for anomaly detection. 
Based on them, we constructed \textit{SemLogCluster} and \textit{SemSVM} by incorporating the lightweight semantic-based log representation method used in \tech{}, respectively.

\begin{table}[t!]
    \centering
    \small
    \caption{\ins{Effectiveness comparison between SVM/LogCluster and \textit{SemSVM}/\textit{SemLogCluster} on all the five used datasets}}

\centering
\begin{tabular}{@{}l|ccc|ccc|ccc|ccc@{}}
\toprule
\multirow{2}{*}{Dataset} & \multicolumn{3}{c|}{\textit{SVM}} & \multicolumn{3}{c|}{\textit{SemSVM}} & \multicolumn{3}{c|}{\textit{LogCluster}} & \multicolumn{3}{c}{\textit{SemLogCluster}} \\
                         & \textit{Prec.}      & \textit{Recall}      & \textit{$F_1$}     & \textit{Prec.}       & \textit{Recall}        & \textit{$F_1$}      & \textit{Prec.}         & \textit{Recall}        & \textit{$F_1$}       & \textit{Prec.}         & \textit{Recall}        & \textit{$F_1$}        \\ \midrule
\HDFS{}                     & 0.999  & 0.662  & 0.796  & 0.999   & 0.974   & 0.986   & 0.997     & 0.900    & 0.946    & 0.963     & 0.985     & 0.972     \\
\BGL{}                      & 0.799  & 0.853  & 0.825  & 0.998   & 0.889   & 0.940    & 0.988     & 0.623    & 0.766    & 0.972    &  0.654    & 0.782    \\
\Spirit{}                   & 0.993  & 0.959  & 0.976  & 0.981   & 0.985   & 0.983   & 0.994     & 0.942    & 0.968    & 0.995     & 0.936     & 0.965    \\
\ZQY{}                       & 0.933  & 1.000      & 0.965  & 0.932   & 0.999   & 0.964   & 0.991     & 0.879    & 0.932    & 0.952    & 0.986     & 0.968     \\
\NC{}                       & 1.000      & 0.643  & 0.783  & 1.000       & 0.823   & 0.903   & 0.999     & 0.671        & 0.802    & 0.996     & 0.925     & 0.959     \\ \midrule
Average                  & 0.944  & 0.823  & 0.869  & 0.982   & 0.934   & 0.955   & 0.993     & 0.803    & 0.883    & 0.976     & 0.897     & 0.929     \\ \bottomrule
\end{tabular}
    \label{tab:imp_svm_cluster} 
\end{table}

Table~\ref{tab:imp_svm_cluster} shows the comparison results between SVM/LogCluster and \textit{SemSVM}/\textit{SemLogCluster} on \del{the HDFS dataset}\ins{all the five used datasets}.
We found that after incorporating semantic-based log representation in both LogCluster and SVM, their effectiveness can be effectively improved.
Specifically, the \ins{average} improvement of \textit{SemLogCluster} over LogCluster is \ins{5.210\%}\del{2.748\%} and that of \textit{SemSVM} over SVM is \ins{9.896\%}\del{3.249\%}, in terms of F1-score, which makes their effectiveness close to the supervised LogRobust and the semi-supervised PLELog \ins{except \textit{SemLogCluster} on the \BGL{} dataset}\del{on\HDFS{}}.
The results demonstrate the generalizability of the conclusions from our study to a large extent and further confirm the great potential of traditional techniques for more practical log-based anomaly detection. 

We manually analyzed the reason about the relatively small improvement of \textit{SemLogCluster} over \textit{LogCluster} on the \BGL{} dataset.
The reason lies in that we directly used the pre-trained word vectors on the Common Crawl Corpus dataset using the Fast-Text toolkit for semantic representation following the existing work~\cite{LogRobust,PLELog}, and 8.922\% of tokens in the \BGL{} dataset are Out-Of-Vocabulary (OOV) with regard to the pre-trained word vectors.
Further, the OOV tokens are uniformly represented by a specialized token (such as ``<oov>''), and thus the distance between log sequences measured by the clustering-based technique can be very small based on the semantic representation on the \BGL{} dataset, which increases the difficulty of precisely clustering them.
Nevertheless, \textit{SemLogCluster} still outperforms \textit{LogCluster}, demonstrating the usefulness of semantic-based log representation for traditional anomaly detection techniques to some degree.
In the future, we can improve the effectiveness of \textit{SemLogCluster} by incorporating more effective semantic-based log representation methods.


Here, we did not study the IM technique (Invariants Mining~\cite{InvariantMining}).
IM mines the invariants (or patterns) existing in log sequences (such as ``open file'' and ``close file'') based on the occurrence frequency of log events and detects an anomaly by determining whether a log sequence violates the invariants.
However, after representing log sequences as the form of semantic vectors, it is hard to mine such invariants due to the incompatibility with the principle underlying IM, and thus it cannot be used for anomaly detection.

\subsection{Case Study}
We further illustrate why \tech{} works well through two examples in this section.

\begin{figure}[t]
    \centering
    \includegraphics[width=\textwidth, bb=0 0 960 120]{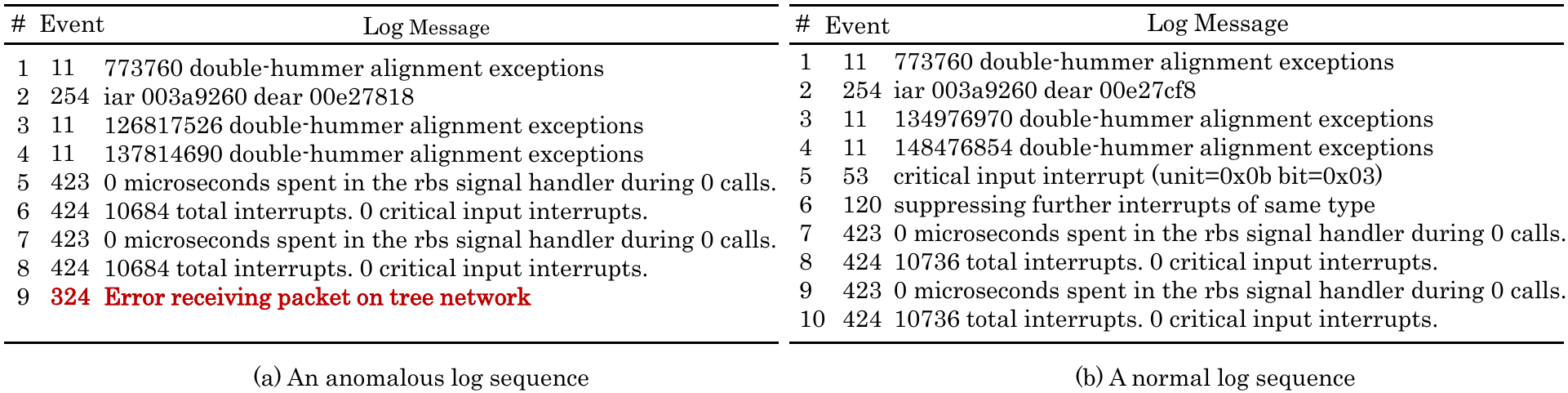}
    \caption{\ins{An anomalous log sequence extracted from \BGL{}. Line 9 of log sequence (a) is labeled by experts as anomalous log message.}}
    \label{fig:case_1}
\end{figure}

\subsubsection{Case 1.}
Figure~\ref{fig:case_1} shows two similar log sequences extracted from the test set of \BGL{}.
The log sequence (a) is anomalous, while the log sequence (b) is normal.
In our experiment, PCA failed to identify the anomalous log sequence (a) as an anomaly, while \tech{} succeeded. 
The reason is that the log events 423 and 424 are unseen in the training set, and thus PCA cannot count them in the event count vector.
Hence, the event count vector for this log sequence is: $[11:3, 254:1, 324:1]$. 
After mapping this vector to the abnormal space, the SPE value of the projection is 17.391, which is smaller than the anomaly detection threshold, and thus PCA mis-reported the anomalous log sequence (a) as a normal one.
In contrast, \tech{} can capture the semantics of the unseen log events 423 and 424 through semantic-based log representation, which help identify this anomaly, demonstrating the contribution of semantic-based log representation for unseen log events.

   
Besides, DeepLog is also affected by unseen log events.
In this example, DeepLog mis-reported the normal log sequence (b) as an anomaly.
As presented in Section~\ref{sec:background}, DeepLog is designed to predict the next log event according to the log events before it. 
When predicting the log event at Line 10, DeepLog takes the first nine lines of log events as input.
Since the log events 423 and 424 are unseen in training data, DeepLog neither represents both of them nor predicts the log event 424, thereby identifying it as an anomaly.
That demonstrates the necessity of relieving the problem of unseen log events.

\begin{figure} [t]
    \centering
    \includegraphics[width=0.65\textwidth, bb=0 0 450 190]{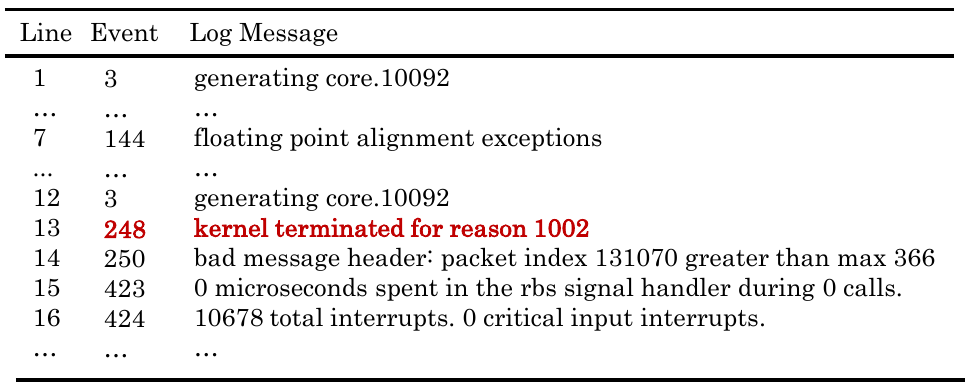}
    \caption{\ins{An anomalous log sequence extracted from \BGL{}. The line 13 is labeled by experts as anomalous log message.}}
    \label{fig:case_3}
\end{figure}

\ins{\subsubsection{Case 2.}
Figure~\ref{fig:case_3} is a part of an anomalous log sequence extracted from the test set of \BGL{}. 
In this log sequence, the log message at Line 13 is labeled as anomalous by experts.
When we used 20\% of training data to build an anomaly detection model through LogRobust in the experiment presented in Section~\ref{sec:rq2}, LogRobust did not identify this anomaly, while LogRobust succeed when using the entire training data to build an anomaly detection model.
The reason for the different results lies in the different distribution of this log event in corresponding training data.
Specifically, the frequency of occurrence of this log message is significantly low in the sampled 20\% of training data, compared to that in the entire training data, and thus LogRobust cannot effectively learn the anomalous information in it based on the insufficient training data.
That demonstrates that LogRobust heavily relies on the quality of training data.
}

\ins{In contrast, \tech{} can identify this log sequence as anomalous regardless of using 20\% of training data or the entire training data in our experiment.
This is because the insight of \tech{} is to detect outliers as anomalies.
That is, as long as the semantics of anomalous log sequences through semantic-based log representation is minority, \tech{} can stably identify these anomalies based on the SPE values in the abnormal space.
This example matches the insight of \tech{} for both sets of training data in the experiment.
That is, even though the frequency of occurrence of this log message is relatively high in the entire training data, the semantics of this log sequence is still minority relative to all log sequences.
}

\subsection{Threats to Validity}
\label{sec:threat}

The threat to \textit{internal} validity mainly lies in our implementation.
In our study, we implemented \tech{} and re-implemented DeepLog, LogRobust, and LogAnomaly.
To reduce this kind of threat, we used some mature libraries such as PyTorch and three authors have carefully checked all the code.
In particular, as presented in Section~\ref{sec:imple}, we also conducted a replication study to confirm the correctness of our re-implementations.
All our code has been released for future replication.


The threat to \textit{external} validity mainly lies in the used datasets.
We used five datasets in our study, including three widely-used public datasets
(i.e., \HDFS{}, \BGL{}, \Spirit{}) and two industrial datasets (i.e., \ZQY{} and \NC{}), which may not represent other datasets.
Actually, our used datasets are diverse by collecting from different systems in different organizations, which can help reduce this kind of threat to some degree.
In the future, we will repeat our experiments on more datasets with greater diversity\ins{, especially for the datasets with more large-scale and high-dimension training data}.


The threat to \textit{construct} validity mainly lies in the hyper-parameter setting for each technique.
To reduce this kind of threat, for each technique on each dataset, we conducted grid search on the corresponding validation set following the practice in the area of DL~\cite{ripley2007pattern}.
In this way, we can try the best to obtain the optimal hyper-parameter setting for each technique on each dataset, which facilitates the fair comparison among the studied techniques.

\ins{Regarding PCA (as well as \tech{}), it has a built-in method (i.e., Q-statistics~\cite{jackson1979control}) to automatically determine the setting of the hyper-parameter (i.e., the threshold of the projection length in abnormal space).
For fair comparison with other techniques, we still applied grid search to both \tech{} and PCA (instead of Q-statistics), and grid search can indeed help find a better hyper-parameter setting than Q-statistics.
Here, we added an experiment to compare grid search and Q-statistics on the basis of \tech{} to show the necessity of grid search.
That is, we evaluated the effectiveness of \tech{} under the hyper-parameter setting determined by Q-statistics (named $\tech{}_{\textit{auto}}$ for ease of presentation) for comparison.
Table~\ref{tab:ablation} shows the comparison results on each dataset.
We found that \tech{} largely outperforms $\tech{}_{\textit{auto}}$ in terms of F1-score, demonstrating the superiority of grid search over the built-in Q-statistics method in PCA.
In particular, the average time spent on the grid search process by \tech{} is just 10.296 seconds across all the datasets, indicating that grid search does not incur too much cost for \tech{}.
}

\begin{table}[t]
  \centering
  \small
  \caption{\ins{Effectiveness comparison among the variants.}}
  \resizebox{\textwidth}{!}{
    \begin{tabular}{@{}l|ccc|ccc|ccc|ccc|ccc@{}}
    \toprule
    \multirow{2}{*}{Technique} & \multicolumn{3}{c|}{{\ul \HDFS{}}}     & \multicolumn{3}{c|}{{\ul \BGL{}}}      & \multicolumn{3}{c|}{{\ul \Spirit{}}}   & \multicolumn{3}{c|}{{\ul \ZQY{}}}       & \multicolumn{3}{c}{{\ul \NC{}}}        \\ 
                               & \textit{Prec.} & \textit{Recall} & $\textit{F}_{1}$ & \textit{Prec.} & \textit{Recall} & $\textit{F}_{1}$ & \textit{Prec.} & \textit{Recall} & $\textit{F}_{1}$ & \textit{Prec.} & \textit{Recall} & $\textit{F}_{1}$ & \textit{Prec.} & \textit{Recall} & $\textit{F}_{1}$ \\ \midrule
    $\tech{}_{auto}$              & 0.775 & 0.458  & 0.576        & 0.339 & 0.456  & 0.389        & 0.322 & 0.991  & 0.486        & 1.000 & 0.862  & 0.926        & 0.998 & 0.846  & 0.916         \\
    \tech{}                     & 0.963 & 1.000  & 0.981        & 0.897 & 1.000  & 0.946        & 0.981 & 0.920  & 0.950        & 0.932 & 1.000  & 0.965        & 0.997 & 0.912  & 0.953        \\ \bottomrule
    \end{tabular}
    }
  \label{tab:ablation}%
\end{table}%


\section{Related Work}
\label{sec:related}

Besides the six most widely-studied techniques (also studied in our work) introduced in Section~\ref{sec:studied_techniques}, there are many other log-based anomaly detection techniques in the literature~\cite{HMM,Improved_HMM,Log3C,InvariantMining,LADCNN,lstm_based_anomaly_detection,LogGAN,Trine,Attention_based_anmomaly_detection,DeepSysLog,CAT,DeepTraLog}.
For example, regarding traditional techniques, 
Yamanishi et al.~\cite{HMM} proposed a Hidden Markov Model to model log sequences. 
Lou et al.~\cite{InvariantMining} proposed to learn invariants in normal log sequences for anomaly detection.
All these existing traditional techniques do not consider the problem of unseen log events.
Regarding DL-based techniques, they tend to design various DL models for anomaly detection.
For example, Lu et al.~\cite{LADCNN} built an Convolutional Neural Network (CNN)for anomaly detection. 
Xia et al~\cite{LogGAN} and Zhao et al.~\cite{Trine} applied Generative Adversarial Network to learn from normal log sequences for anomaly detection.
Zhou et al.~\cite{DeepSysLog} proposed to use numerical values in log messages for anomaly detection.
Zhang et al.~\cite{CAT} proposed a content-aware Transformer to represent log events and applied a Transformer-based decoder to perform anomaly detection.
Xie et al.~\cite{LogGD} proposed to use the Graph Transformer network to combine both graph structure and log semantics to detect anomalies. Zhang et al.~\cite{DeepTraLog} proposed a Graph-based Deep Learning framework to detect anomalies in micro-service system. 

Different from them, our work aims to investigate whether the existing traditional techniques through some optimizations can achieve comparable effectiveness with advanced DL-based techniques in log-based anomaly detection.
Our results show that by incorporating lightweight semantic-based log representation, the optimized version of the PCA technique (i.e., \tech{}) can help achieve more practical log-based anomaly detection. 
Moreover, the generalizability of our conclusions on other traditional techniques has been investigated in Section~\ref{sec:other}. 
Please note that \tech{} is different from the dimension reduction algorithm (i.e., FastICA) used in PLELog, since the former aims to detect anomalies through outlier detection in the projected abnormal space while the latter aims to speed up the clustering process through dimension reduction.

Since our work is an empirical study on log-based anomaly detection, the existing empirical studies in this area are also relevant~\cite{DBLP:conf/issre/HeZHL16,fse21nengwen,DBLP:journals/csur/HeHCYSL21,chen21experiencereport,le21howfar,MSRAStudyFSE22}.
For example, He et al.~\cite{DBLP:conf/issre/HeZHL16} conducted an empirical study to compare the effectiveness of traditional ML-based log-based anomaly detection techniques. Furthermore, He et al.~\cite{MSRAStudyFSE22} conducted an empirical study to investigate the real significant challenges of log analysis techniques in the industry(i.e., Microsoft) and summarized several possible Opportunities including incresing the efficiency of log-based anomaly detection.
Similarly, Le et al.~\cite{le21howfar} conducted an empirical study to investigate some DL-based techniques.
Chen et al.~\cite{chen21experiencereport} empirically investigated the effectiveness of state-of-the-art DL-based techniques, 
and summarized several challenges about their practical use (which can also confirm the necessity of our work).
Zhao et al.~\cite{fse21nengwen} empirically investigated various types of anomalies in real-world log data.
Different from them, our study aims to explore the potential of traditional techniques for more practical anomaly detection following the idea of ``try-with-simpler''.
In our study, we compared the optimized version of the PCA technique (i.e., \tech{}) with four widely-studied DL-based techniques and two widely-studied traditional techniques in terms of effectiveness, efficiency, and stability.

In addition, the potential of traditional ML algorithms has been evaluated and demonstrated in the tasks of linking questions posted on Stack Overflow~\cite{menzies_easy_over_hard} and linguistic smell detection~\cite{linguisticpatternsaner21}, which also compare the traditional ML algorithms with state-of-the-art DL-based techniques.
Different from them, we are the first to explore and demonstrate the potential of traditional techniques in log-based anomaly detection. 
Due to different characteristics of different tasks, the adopted optimizations for the studied traditional techniques are completely different.
Our work further confirms the idea of ``try-with-simpler''.
\section{Conclusion}
\label{sec:con}

In this work, we conducted an extensive study, as the first attempt, to explore the potential of traditional techniques for more practical log-based anomaly detection by optimizing the traditional PCA technique with a lightweight semantic-based log representation method.
We call the optimized version of the PCA technique \tech{}.
Such adaptation on the PCA technique aims to address the problem of unseen log events and thus improve the anomaly detection effectiveness.
The results on five datasets show that the unsupervised \tech{} achieves comparable effectiveness with the supervised/semi-supervised DL-based techniques while is more efficient, more stable, and less dependent on training data. 
Our results confirm that the traditional technique can still excel in log-based anomaly detection after simple but useful adaptation, further confirming the idea of ``try-with-simpler''.

\begin{acks}
We gratefully acknowledge the financial support provided by the National Natural Science Foundation of China Grant Nos. 62322208, 62232001, and CCF Young Elite Scientists Sponsorship Program (by CAST) which made this research possible.
We would like to express our sincere appreciation to the anonymous reviewers and the editors for their constructive suggestions, which have greatly improved the quality of this paper.
\end{acks}

\bibliographystyle{ACM-Reference-Format}
\bibliography{ref}

\end{document}